\documentclass[accepted]{uai2022}

\usepackage[american]{babel}

\usepackage{natbib} 
\usepackage{mathtools} 
\usepackage{booktabs} 
\usepackage{tikz} 
\usepackage{CJK}

\usepackage{algorithm}
\usepackage{algorithmicx}
\usepackage{algpseudocode}
\usepackage{amsmath}
\usepackage{indentfirst}
\usepackage{subfigure}
\usepackage{graphicx} 
\usepackage{stfloats}
\usepackage{amssymb}
\usepackage{amsthm}
\newtheorem{theorem}{Theorem}
\usepackage{multirow}
\graphicspath{{./figures/}}

\title{AACC: Asymmetric Actor-Critic in Contextual Reinforcement Learning}

\author[1, *]{Wangyang Yue}
\author[1, *]{Yuan Zhou}
\author[1]{Xiaochuan Zhang}
\author[1]{Yuchen Hua}
\author[1]{Zhiyuan Wang}
\author[1]{Guang Kou}

\affil[1]{%
    Artificial Intelligence Research Center \\
    DII, Beijing, China
}
\affil[*]{%
    equal contribution
}

\begin{document}
\maketitle

\begin{abstract}
Reinforcement Learning~(RL) techniques have drawn great attention in many challenging tasks, but their performance deteriorates dramatically when applied to real-world problems. 
Various methods, such as domain randomization, have been proposed to deal with such situations by training agents under different environmental setups, and therefore they can be generalized to different environments during deployment. 
However, they usually do not incorporate the underlying environmental factor information that the agents interact with properly and thus can be overly conservative when facing changes in the surroundings. 
In this paper, we first formalize the task of adapting to changing environmental dynamics in RL as a generalization problem using Contextual Markov Decision Processes (CMDPs).  
We then propose the \emph{Asymmetric Actor-Critic in Contextual RL}~(AACC) as an end-to-end actor-critic method to deal with such generalization tasks. 
We demonstrate the essential improvements in the performance of AACC over existing baselines experimentally in a range of simulated environments. 
\end{abstract}

\section{Introduction}\label{sec:intro}
Reinforcement Learning (RL) has proven to be powerful in a range of tasks such as classic games~\citep{silver2016mastering}, 
video games~\citep{mnih2015human, 2019The}, autonomous vehicles~\citep{2020Can} and robotics~\citep{akkaya2019solving}.
Especially in the scenarios where the deploying environment is exactly the same as the training one such as in Go~\citep{silver2016mastering} and StarCraft II~\citep{2019The}, RL techniques have even outperformed humans for orders of magnitude. 
However, when it comes to more complicated real-world applications such as Unmanned Aerial Vehicles (UAVs) flight control~\citep{2021Drone}, legged robots~\citep{kumar2021rma} and other classic control scenarios~\citep{benjamins2021carl}, RL techniques have not yet witnessed substantial progress. 
For example, we are interested in a UAV
that needs to learn to maintain a steady direction in the case of changing wind. 
Or, a legged robot tries to adapt to different territories and ground frictions to avoid falling when walking in unseen places. 
In such situations, one wants to construct a policy that can generalize to 
the deployment environment in which one does not have the entire information and may vary largely from the one that the agent is trained on. 

We classify such situation as the \emph{generalization} problem~\citep{kirk2021survey} in RL. 
In particular, we focus on the cases where the deploying environment may differ from the training environment but share the same distribution of underlying parameters.
One key reason why traditional RL methods suffer easily
lies in the fact that it is almost impossible to simulate the exact dynamics of the real-world completely,
and it is also unaffordable to train the policy in real-world environments directly for the majority real-world tasks~\citep{akkaya2019solving, lee2020learning}. 
Therefore, the policy trained in the simulation is difficult to generalize to the real environment and may perform unexpectedly even when facing small mismatches.  
Even worse, the environmental dynamics of the real world change frequently, which makes the adaptation even harder to achieve. 

There are two popular classes of approaches to deal with this problem. 
One class is to construct a robust policy that can perform under different environments using domain randomization techniques during training~\citep{2017Domain}, even though the policy may not have the exact knowledge of the environmental information. 
Such policy is widely applicable but at the same time can be overly conservative in terms of performance due to the same consideration. 
The other class usually trains an environment-specific policy in the simulation that takes environmental parameters as additional inputs, and uses the trained policy to build a domain inference module with history information and condition the policy on the inferred environment information~\citep{yu2017preparing, lee2020learning, kumar2021rma}. 
Though making inroads in adapting to the varieties during deployment, these approaches still face challenges such as  
estimating the environmental parameters or their distributions based on a small number of observations can be inaccurate. 
Moreover, there is a counter-intuitive phenomenon that policies that have access to true environmental information may not perform better than policies that do not make decisions based on environmental information; we shall demonstrate such in Section~\ref{Comparative Evaluation}. 

Witnessing the shortcomings of existing baselines, we propose an asymmetric actor-critic method to handle the generalization problem in RL. 
The \emph{asymmetric} design is motivated by \citet{pinto2017asymmetric} where the policy learns to act with rendered scene images as inputs. It is refined gradually with full state information from the simulator during training and consequently can be generalized to test environments in a zero-shot manner, i.e. without further training on new data from deployment. 
In this paper, we call our method \emph{Asymmetric Actor-Critic in Contextual Reinforcement Learning (AACC)}, where a policy is trained by providing observations and extra environmental parameters, and is expected to act in varying deployment setups with observations only.
The generalization problem is formalized using Contextual Markov Decision Process (CMDP)~\citep{hallak2015contextual}, in which different environment parameters refer to different contexts, and provide theoretical derivations of the corresponding policy gradient in AACC.
Note that unlike~\citet{benjamins2021carl}, we assume multiple environment parameters may vary rather than individual ones. 
We evaluate the effectiveness of AACC empirically in the six different environments with various environment parameters and discuss how such information may have a negative effect on algorithm performance when used improperly. 

\section{Related Work}
\vspace{-5pt}
\paragraph{Generalization in RL.}
Generalization in RL is about how well a policy can generalize to unseen scenarios at deployment time. Methods for generalization in RL are mainly split into two categories: one to increase the similarity between training and testing environments, the other to explicitly deal with differences between training and testing environments. To increase the similarity, a common approach is to use domain randomization which varies the training environmental parameters to cover the testing environments with a distribution of the
environments \citep{2017Domain, peng2018sim}.
In particular, \citet{akkaya2019solving} uses automatic domain randomization by dynamically adjusting the distribution based on the current performance of the policies. 
The latter category is to explicitly handle possible differences between the features of training and testing environments. 
\citet{2021Decoupling} decouples policy and value optimization, and also uses an auxiliary loss to constrain the policy representation to be invariant to task instance. \citet{2021Unsupervised} uses unsupervised visual attention to encode information relevant to the foreground of the visual images. \citet{2020Dynamics} introduces an annealing-based optimization method to achieve better generalization by introducing an information bottleneck.

\vspace{-8pt}
\paragraph{Sim-to-Real Transfer.}
As a subclass of generalization in RL, sim-to-real transfer is a concrete application of generalization problems. Simulation-based training provides low-cost data but involves inherent mismatches with real-world scenarios. To bridge the gap between simulation and reality, \citet{2018Sim} learns robust controllers in a physics simulator by using domain randomization with added perturbations and designing a compact observation space.
\citet{2019Learning} uses a neural network to parameterize the actuator model to make the simulation more accurate, but the approach requires initial data collection from the robot.
\citet{E2016Towards} reduces the sim-to-real gap from a computer vision perspective by using domain adaptation, which is a subset of transfer learning methods.
\citet{yu2017preparing} proposes a UP-OSI framework, which belongs to the online adaptation method. The process is to first train a universal policy (UP) in simulation, and then use the trained policy to train an online system identification (OSI) module with history information to estimate environment parameters. However, explicit inference of every environmental parameter is extremely challenging when there are more environment parameters. To deal with the difficulty of inference, \citet{lee2020learning, kumar2021rma, miki2022learning} introduce a module, called environmental factor encoder, to encode environment parameters to a lower dimension.

\vspace{-8pt}
\paragraph{Asymmetric Actor-Critic.}
Training in simulation is, in principle, able to exploit global information to achieve better online performance. 
This advantage can be exploited as long as the deployed policy does not use global information, and actor-critic methods \citep{konda2000actor} just meet this condition via critic asymmetry where actor and critic receive different information. 
Prior works based on asymmetric actor-critic are mainly applied to POMDPs, which focus on real-world robot control problems and multi-agent systems. 
In robot control, agents lack access to the system state, so \citet{pinto2017asymmetric} uses a training algorithm in which the critic is trained on full states in simulation while the actor gets rendered images as input. And during the deployment phase, agents no longer need full states because they no longer use the critic. The framework is called offline learning and online execution. 
As for the scenarios of multi-agent systems, each agent only gets local observation due to the limitation of the event horizon. \citet{rashid2018qmix} uses global information to train a network that estimates joint action-values of each agent that conditions only on local observation. During the test phase, policies no longer need the network. The paradigm is called centralized training with decentralized execution. Our work is inspired by these methods that we can take advantage of the information provided by simulation which determines the environmental dynamics in a specific environment while being unavailable at deployment time.

\section{Preliminaries}

\subsection{Reinforcement Learning}
The standard formalism in RL is the Markov Decision Processes (MDPs), which can be represented as the tuple $M = (S, A, R, P, \rho_0, \gamma)$, where $S$ is the state space; $A$ is the action space; $R: S \times A \rightarrow \mathbb{R}$ is the scalar reward function; $P(s'|s, a)$ is the transition probabilities that specify the distribution on the next state given the current state and an action; $\rho_0$ is the initial state distribution; $\gamma \in (0, 1)$ is the discount factor. Let $\pi$ denote a stochastic policy $\pi: S \times A \rightarrow [0, 1]$, the Q value of a particular state under policy $\pi$ is the expected total discounted return from  a particular action at the state, \emph{i.e.},
\begin{equation}
    Q_{\pi}(s, a)=\mathbb{E}( \sum_{k=1}^{\infty}\gamma^{k-1}r_{t+k}|s_t=s,a_t=a, \pi).
\end{equation}
we also define the value of state under policy $\pi$, \emph{i.e.},
\begin{equation}
\begin{aligned}
    V_{\pi}(s) &= \mathbb{E}(\sum_{k=1}^{\infty}\gamma^{k-1}r_{t+k}|s_t=s,\pi) \\
    &= \sum_{a \in A}\pi(a|s)Q_{\pi}(s,a).
\end{aligned}
\end{equation}
The advantage function is the difference between the Q value and the value function, \emph{i.e.},
\begin{equation}
    A_{\pi}(s,a)=V_{\pi}(s)-Q_{\pi}(s, a).
\end{equation}
The goal of RL is to obtain a policy which maximises the cumulative discounted reward from the start state, denoted by the performance objective,
\begin{equation}
    J(\pi) = \mathbb{E}(\sum_{t=1}^{\infty}\gamma^{t-1}r_t|s_0,\pi) \\
    = V_{\pi}(s_0),
\end{equation}
assuming that every episode starts from a particular state $s_0$.
The policy gradient theorem \citep{sutton2000policy} states that the gradient of $J(\pi)$ with respect to the parameters of the policy is given by
\begin{equation}\label{gradient}
    \nabla J(\pi)=\mathbb{E}[(Q_{\pi}(s, a)\nabla\log\pi(s,a)].
\end{equation}

\subsection{Contextual MDPs}
As \citet{hallak2015contextual} refers, CMDPs are a special form of Partially Observable MDPs (POMDPs).
A CMDP is an MDP in which the state can be decomposed into a tuple $s=(c, o)\in S$, where $o \in O$ is the underlying state or observation in a POMDP, and $c\in C$ is the context. 
Generally, it is assumed that the context is not observed by the agent, making the CMDP a POMDP with observation space $O$ and an emission function that discards the context information: $\phi((c, o)):=o$. Note that generally, the context does not change within an episode, only between episodes. 

The context determines dynamics, reward function (tasks), and observation function in environments. We focus on dynamics that are determined by the context because we do not care about different tasks or vision-based sense. For example, we train a UAV agent to keep a smooth flight whether there is wind or not. In this case, the context refers to unknown physical parameters such as wind velocity and direction.
Conditioned on the context $c$, $(o_t, a_t)$ will be mapped to $o_{t+1}$ with transition probability $P(o_{t+1}| o_t, a_t, c)$.
\citet{benjamins2021carl} splits the class of context into the explicit context that is directly available from the environment, and the implicit context that the abstract information is hidden in an available state. In our work, however, the context is hidden, which corresponds to the properties in real-world scenarios, such as gravity or friction.
Due to the nature of \emph{Controllable} environments \citep{kirk2021survey} that provide direct control over the factors of variation between contexts, we can get the context during the training process. 
This is the key idea of our method that we can directly use the context during training,  rather than testing.

\vspace{-5pt}
\paragraph{Domain Randomization.}
Domain randomization has been shown to be a simple but powerful technique in robot control tasks \citep{pinto2017asymmetric,2017Domain,2018Learning}.  
It randomizes the dynamics of the underlying environment in the training phase instead of training the agents in a single simulated environment. 
In our work, to compare the performance of different methods under the same condition, we assume that the context set used in all experiments follows the same distribution. 

\section{Method}
Our asymmetric actor-critic in contextual RL method can be derived starting from policy gradient in AACC. We will first present this derivation, verify that we can replace the original value function with asymmetric critic, then present a practical method based on an RL algorithm in CMDPs.

\subsection{From Symmetric  to Asymmetric}
In POMDPs, policy gradient methods \citep{sutton2000policy} for fully observable problems is adapted to partial observable problems by replacing the system state $s$ with the history state $h$ \citep{2021Unbiased}. However, they pointed out that using the state value function $V_{\pi}(s)$ to replace the history value function $V_{\pi}(h)$ in the asymmetric critic can be intrinsically questionable. The reason is that the state value function $V_{\pi}(s)$, as a estimator of the history value function $V_{\pi}(h)$, always has a corresponding estimation bias, \emph{i.e.}, the difference between $\mathbb{E}_{s|h}[V_{\pi}(s)]$ and $V_{\pi}(h)$ always exists.

CMDPs do not suffer from the issue mentioned above. 
Agents are \emph{reactive} which means that policies are determined using the last observation rather than the entire history~\citep{2021Unbiased}.
Furthermore, the context vector is fixed for each episode, i.e., being time-invariant. 
Therefore, we use $c \in C$ to represent a specific context $c$ which follows a distribution $p(c)$, and use the asymmetric critic context-observation $V_{\pi}(c,o)$ to represent corresponding state value function.

Consider a context-based Q value $Q_{\pi}(c,o,a)$, which represents the expected discounted return with a particular action $a$, a context $c$ and observation $o$ under the policy $\pi(a|c,o)$.
\begin{equation}\label{qfunc}
    Q_{\pi}(c,o,a)=\mathbb{E}(\sum_{k=1}^{\infty}\gamma^{k-1}r_{t+k}|o_t=o,a_t=a,c,\pi).
\end{equation}
Using the Bellman equation, we rewrite Equation~\ref{qfunc} as:
\begin{equation}
    Q_{\pi}(c,o,a) = R(c,o,a)+\gamma\mathbb{E}[V(c,o')],
\end{equation}
where 
\begin{equation}
    V(c,o')=\mathbb{E}_{a \sim \pi}[Q_{\pi}(c,o,a)]
\end{equation}
is the value function, and $o'$ is the following observation with action $a$.

\begin{theorem}\label{theorem1}
For any given CMDP and policy $\pi$, $V_{\pi}(o)$ is an unbiased estimator of $V_{\pi}(c,o)$, i.e., $V_{\pi}(o)=\mathbb{E}_{c \sim p(c)}[V_{\pi}(c,o)]$, assuming that the policy is determined by the observation $o$, i.e., $\pi(c,o)=\pi(o)$.
\end{theorem}
\vspace{-10pt}
\begin{proof}
see Appendix~\ref{appe:proof}.
\end{proof}

Following the discussion in \citep{kirk2021survey}, we ignore the difference between the distribution of contexts in the training set and the testing set, and assume that the context $c$ does not affect the initial underlying state or the observation. In this case, the objective to be optimized is modified  as the expectation of the cumulative discounted reward as follows:
\begin{equation}
    J(\pi) \triangleq \mathbb{E}_{c \sim p(c)}\left[\mathbb{E}(\sum_{t=1}^{\infty}\gamma^{t-1}r_t|c, o_0,\pi)\right],
\end{equation}
where $p(c)$ is the distribution function of the context,  which only depends on the  simulation of the real-world scenario.
Following Theorem~\ref{theorem1}, we can simplify the objective as:
\begin{equation}
\begin{aligned}
    J(\pi) = \mathbb{E}_{c \sim p(c)}[V_{\pi}(c,o_0)] = V_{\pi}(o_0),
\end{aligned}
\end{equation}which has the same form of a standard RL objective,  with only  the state of the input been replaced with the observation. Therefore, we compute the gradient of the objective as follows.

\begin{theorem}[\rm{Policy Gradient in AACC}] 
Assume that  the context $c$ follows a distribution $p(c)$ at the beginning of an episode, and remains fixed within the same episode, then we have:
\begin{equation}
    \nabla J(\pi) = \mathbb{E}_{c \sim p(c)}\left[\mathbb{E}[Q_{\pi}(c,o,a)\nabla \log\pi(a|o)]\right].
    \vspace{-20pt}
\end{equation}
\label{theorem2}
\end{theorem}
\vspace{-10pt}
\begin{proof}
see  Appendix~\ref{appe:proof}.
\end{proof}

Assuming that the context in current episode is $c$, the gradient of the objective is given by
\begin{equation}
    \nabla J^{c}(\pi)=\mathbb{E}[Q_{\pi}(c,o,a)\nabla \log\pi(a|o)].
\end{equation}

Such gradient is similar to a standard policy gradient, in which the actor and the critic have the same state as their input. We extend their results to the case that the critic has an extra context input, and the theorem shows that we can use existing RL algorithms to AACC with only a little change. 

\begin{figure}[t]
  \centering
  \includegraphics[width=8cm]{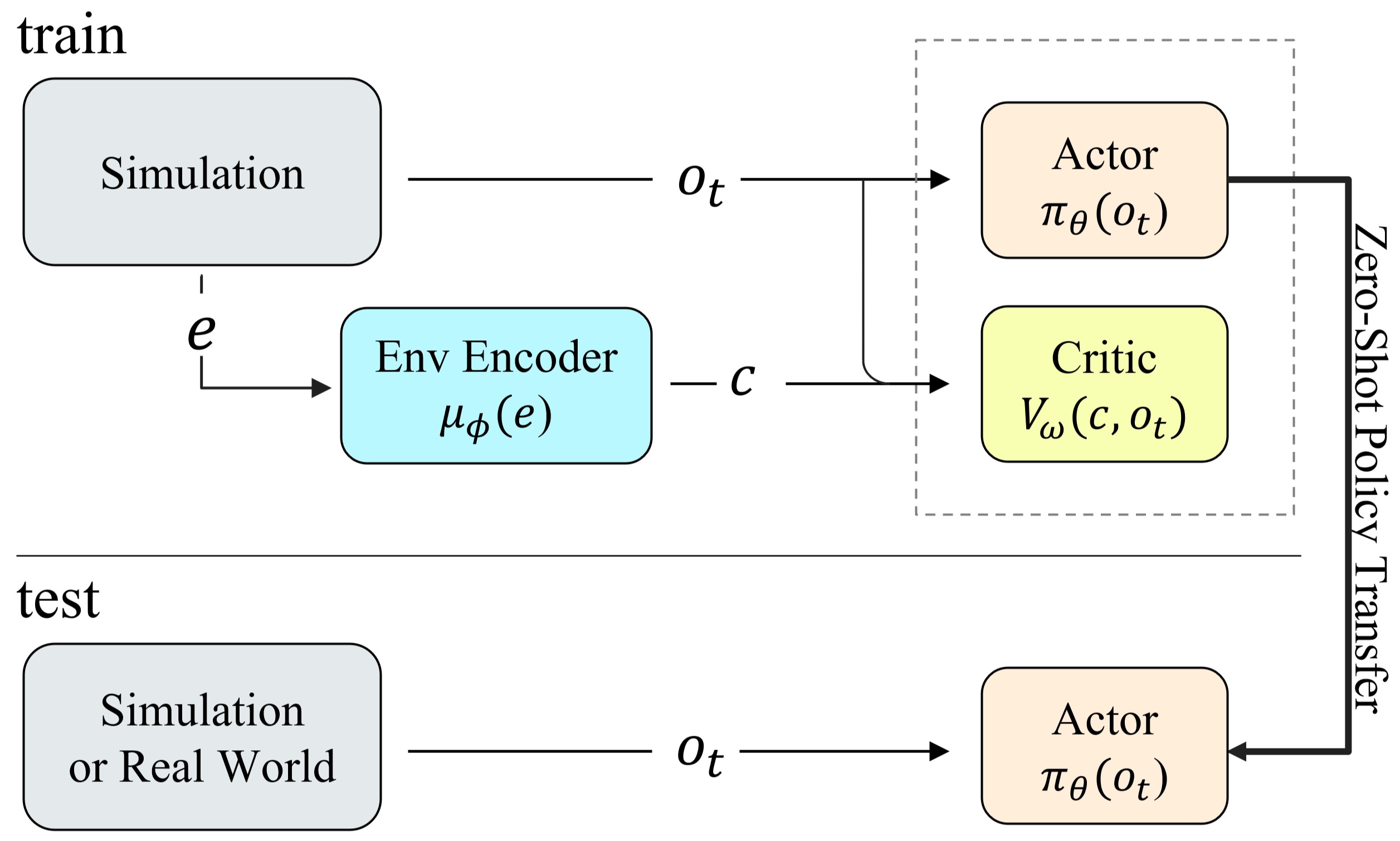}
  \caption{\textbf{Overview of AACC.} During the training phase, the actor takes the current observation $o_t$ as input, while the critic takes the observation $o_t$ as well as a context vector $c$ as input. The context vector $c$ is obtained by encoding environmental factors $e$ through the environmental factor encoder $\mu$. So the input vector $s_{t,c}$ for the critic network is a combination of $o_t$ and $c$: $s_{t, c}=\left(\mu(e), o_t \right)=(c, o_t)$. During testing, the agent only use the actor to make decision by taking the observation $o_t$ as input.}
  \label{fig:AACC framework}
  \vspace{-5pt}
\end{figure}

\subsection{Asymmetric Actor-Critic in Contextual RL} \label{sec:AACC}
As discussed above, asymmetric actor-critic can be applied to existing RL algorithms when the environmental factors in simulation are available.
In this section, we describe the details of AACC, whose main framework is in Figure~\ref{fig:AACC framework}.  

Essentially, our method builds on the actor-critic architecture~\citep{konda2000actor}, which uses full state $s \in S$ that includes the context and observation to train the critic, while using observation $o \in O$ to train the actor. 
This is inspired by~\citet{pinto2017asymmetric, 2018Learning} which uses rendered images as partial observation and the full state information to train the actor and critic respectively; we take the advantage of the controllable environments \citep{kirk2021survey} which contains various factors that are controlled by the user of the environments. 
The factors are called \emph{privileged information}~\citep{chen2020learning} or \emph{environmental factors} $e$, and can be used to match the difference between simulation and real-world scenario.
Environmental factors, to some extent, are the context in CMDPs that can be taken as the context directly or encoded through a neural network before serving as the context.

In our method, environmental factors are encoded into the context vector $c$ using an environmental factor encoder (Env Encoder) $\mu(e)$ which means we apply an encoder, which is an MLP, to these factors rather than using them directly as contexts inputs in the value function.
This is inspired by \citet{kumar2021rma} which points out that the role of the environmental factor encoder is to extract essential information about environmental factors. According to them, we keep environmental factor encoder in our method. Note that their method is the symmetric actor-critic method by using environmental factor encoder, while our method just uses the module in the critic. The input vector $s_{t,c}$ 
for the critic network is defined as a combination of $c$ and $o_t$:
\begin{equation}
    s_{t, c}=\left(\mu(e), o_t \right)=(c, o_t).
\end{equation}

In this paper, we use Proximal Policy Optimization (PPO) \citep{schulman2017proximal} as the actor-critic algorithm, which is often used in the field of robotic control \citep{kumar2021rma, miki2022learning}. In particular, our method can be extended to other actor-critic-based learning methods, such as deep deterministic policy gradient \citep{2015Continuous} and soft actor-critic \citep{2018Soft}, as long as the assumption of observation in Theorem~\ref{theorem1} is satisfied.

Following the common process in the papers mentioned above, we use function approximators for both the actor and the critic. We consider a parameterized state value function (critic) $V_w(c,o_t)$, policy (actor) $\pi_{\theta}(a_t|o_t)$ and the environmental factor encoder $\mu_{\phi}(e)$. The parameters of these networks are $w$, $\theta$ and $\phi$.
The actor network $\pi_{\theta}(a_t|o_t)$ is obtained mainly by updating the following objective
\begin{equation}\label{actor objective}
    J_{\pi}(\theta)=\mathbb{\hat{E}}_t\left[\min(r_t({\theta})\hat{A_t}, \text{clip} (r_t({\theta}), 1-\epsilon, 1+\epsilon)\hat{A_t})\right],
\end{equation}
where $r_t({\theta})={\pi_{\theta}(a_t|o_t)}/{\pi_{\theta_{old}}(a_t|o_t)}$ denotes the probability ratio,
$\hat{A_t}$ is an estimator of the advantage function at timestep $t$, $\epsilon$ is a hyperparameter, $\theta_{old}$ is the policy parameter before the update.

The critic network is trained by optimizing the following loss:
\begin{equation}\label{critic objective}
    L_V(w)=\mathbb{\hat{E}}_t\left[(V_w(c,o_t)-\hat{V}_t)^2\right],
\end{equation}
where $\hat{V}_t$ is the total discounted reward obtained after timestep $t$ in the corresponding episode :
\begin{equation}
    \hat{V}_t=\sum_{k=t}^T \gamma^{k-t}r_k.
\end{equation}

\begin{algorithm}[tbp]
    \caption{AACC using PPO} \label{alg: AACC}
    \begin{algorithmic}[1]
        \State Initialize actor and critic network parameters $\theta$ and $w$;
        \State Initialize env encoder parameters $\phi$;
        \State Empty replay buffer $D$;
        \For{$iterative=1, M$}
            \For{$episode=1, N$}
                \State Sample a vector $e$ in the distribution $p(e)$
                \For{$t=0,T-1$}
                \State $a_t \sim \pi_{\theta}(a_t|o_t)$
                \State $o_{t+1} \sim p_{e}(o_{t+1}|o_t,a_t)$
                \State $D \leftarrow D \cup \{(o_t, a_t, r(o_t, a_t, e), o_{t+1}, e\}$
                \EndFor
            \EndFor
            \For{$i=1,K$}
                \State $c \leftarrow \mu(e)$
                \State Update actor network using $\{o_t, a_t, r_t, o_{t+1}\}$
                \State Update critic network and env encoder using $\{o_t, a_t, r_t, o_{t+1}, c\}$
            \EndFor
            \State Empty $D$
        \EndFor
    \end{algorithmic}
\end{algorithm}

To optimize the objectives mentioned above, the gradient is required to be given explicitly, however, for the critic network, the input contains the environmental factor encoder $\mu_{\phi}(e)$, which is part of the state $s_{t, c}$, rather than only the observation $o$,
where the parameters $\phi$ is the environmental factor encoder, which is also updated together with the critic network, as shown in Algorithm \ref{alg: AACC}. Fortunately, with the assumption that the policy does not depend on $c$, we can apply the result of Theorem~\ref{theorem2}, and compute the gradient of both Equation~\ref{actor objective} and Equation~\ref{critic objective}. Since the PPO method works on both the environments of discrete action spaces and continuous action spaces, to verify our algorithm,  we will test our method in these two types of environments.

\section{Experiment}
The goal of our experiments evaluation is to demonstrate how AACC allows agents to generalize better to test environments and exploit how environmental factors should be used and what benefits or harm environmental factors would bring to RL algorithms compared with prior RL algorithms. 

\paragraph{Environments:}
Based on the above goal, we compare our method to prior techniques and some variants on a range of challenging control tasks with appropriate environmental factors from Gym-JSBSim for the control of fixed-wing UAVs using the JSBSim \citep{2004JSBSim} flight dynamics model, 6-DoF Hopper and 18-DoF Unitree Laikago quadruped, and contextually extend classic control environments \citep{benjamins2021carl} from OpenAI Gym \citep{2016OpenAI}. 
For Gym-JSBSim, the goal is to control a UAV agent to fly steady trajectories along a specified compass direction while maintaining a constant altitude. Wind speed and wind direction of the environmental factors are varied. 
The goal of Hopper and Laikago is to move as fast as possible (Laikago sets a maximum velocity value) in the forward direction without falling. 
The environmental factors include ground friction, spinning friction, inertia, and payload. 
Contextually extended classic control environments include CartPole, Pendulum, and Acrobot.
The difficulty of a task is determined by two things: (a) the task itself, for instance, the degree of sparsity of the reward function, the dimensions of action and state; (b) the dynamics changes caused by the environmental factors.
Other details can be found in Appendix~\ref{appe:environment description}. 
\paragraph{Baselines:} We set up different baselines for three reasons:
(a) to show that AACC method has improved the performance;
(b) to show whether adding the environmental factors to the agent improves the performance, and how environmental factors affect the actor and the critic respectively;
(c) to investigate whether adding Env Encoder for the environmental factors would be better than using the information directly.
Note that for the goal of exploiting what benefits or harm that environmental factors would bring to RL algorithms and do a fair comparison to AACC, we only do the training in the first phase of methods that need history information to estimate environmental factors, including SysID, RMA, RMA-normal, AACC-actor and AACC-hybrid. It means we assume that the environmental factors are accessible to these methods during evaluation.
As for the architectures of the baselines, the details are shown in Appendix~\ref{variant}. As for the hyperparameters, AACC and all the baselines share the same hyperparameters to ensure fairness as shown in Appendix~\ref{hyperparameters}.
\begin{itemize}
  \item Robustness through Domain Randomization (Robust): The policy is trained without environmental factors as inputs and is robust to the variations in training environments \citep{peng2018sim}.
  \item System Identification (SysID): This baseline is implemented based on \citet{yu2017preparing} that trains a universal policy in the first phase by directly getting environmental factors in simulation.
  \item Rapid Motor Adaptation (RMA): This method is based on \citet{kumar2021rma}. As above, we only train the policy in the first phase. Note that RMA puts the last time step action $a_{t-1}$ into the state $s_t$.
  \item Rapid Motor Adaptation (RMA-normal): To keep the input uniform, RMA-normal removes the last time step action $a_{t-1}$ from inputting to state $s_t$ as per RMA.
  \item Inverse Asymmetric Actor-Critic (AACC-actor): To show how the environmental factors affect the algorithm, we replace the environmental factor encoder from the critic network with the actor network. The goal is to evaluate the impact of adding environmental factors into the actor network on performance.
  \item Hybrid Asymmetric Actor-Critic (AACC-hybrid): This baseline is the mix of AACC and AACC-actor that the actor and critic network have different environmental factor encoders respectively. 
\end{itemize}

\begin{figure*}[t]
\centering
\subfigure[Acrobot]{
\includegraphics[width=5cm]{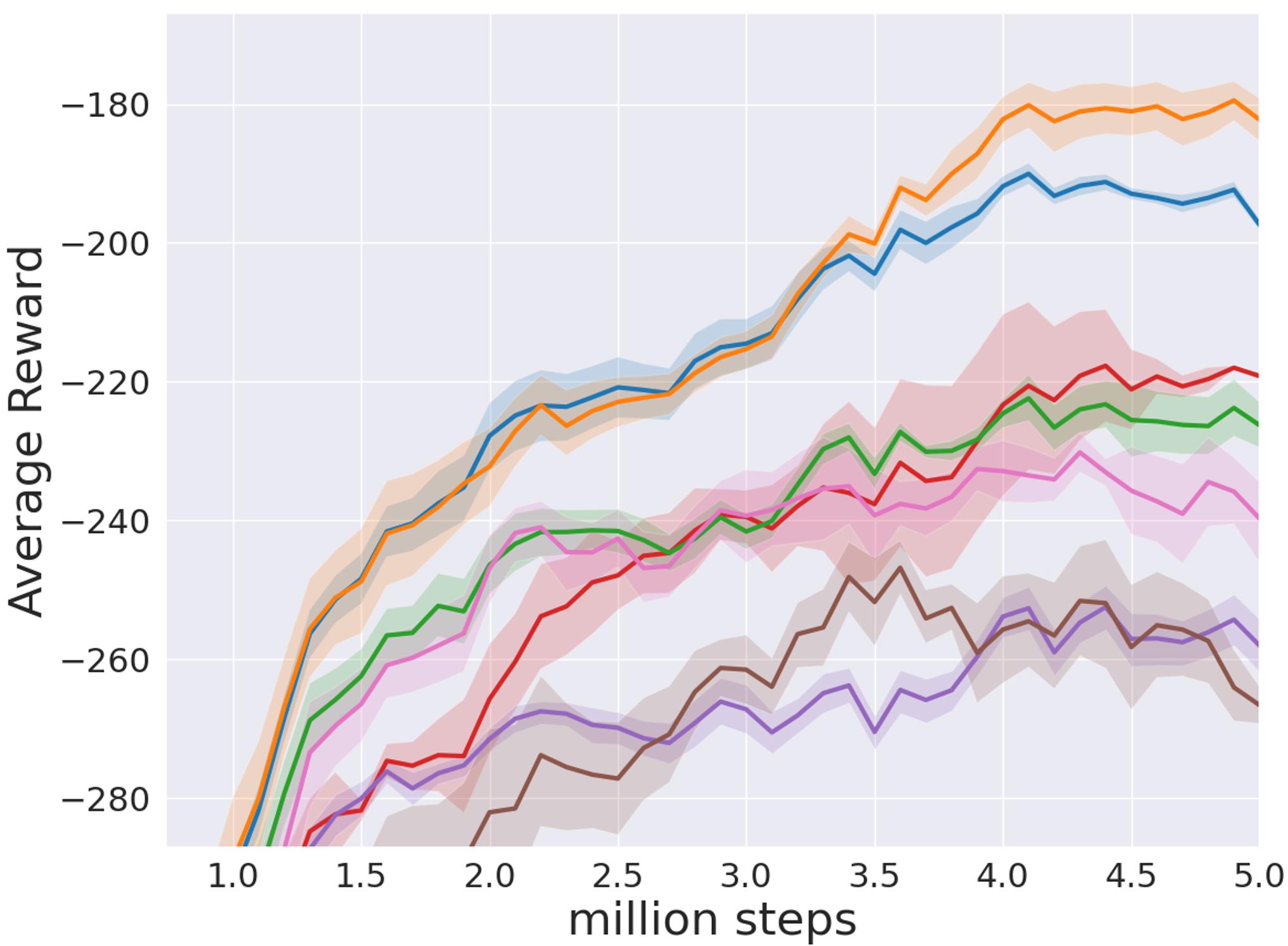}
}
\quad
\subfigure[CartPole]{
\includegraphics[width=5cm]{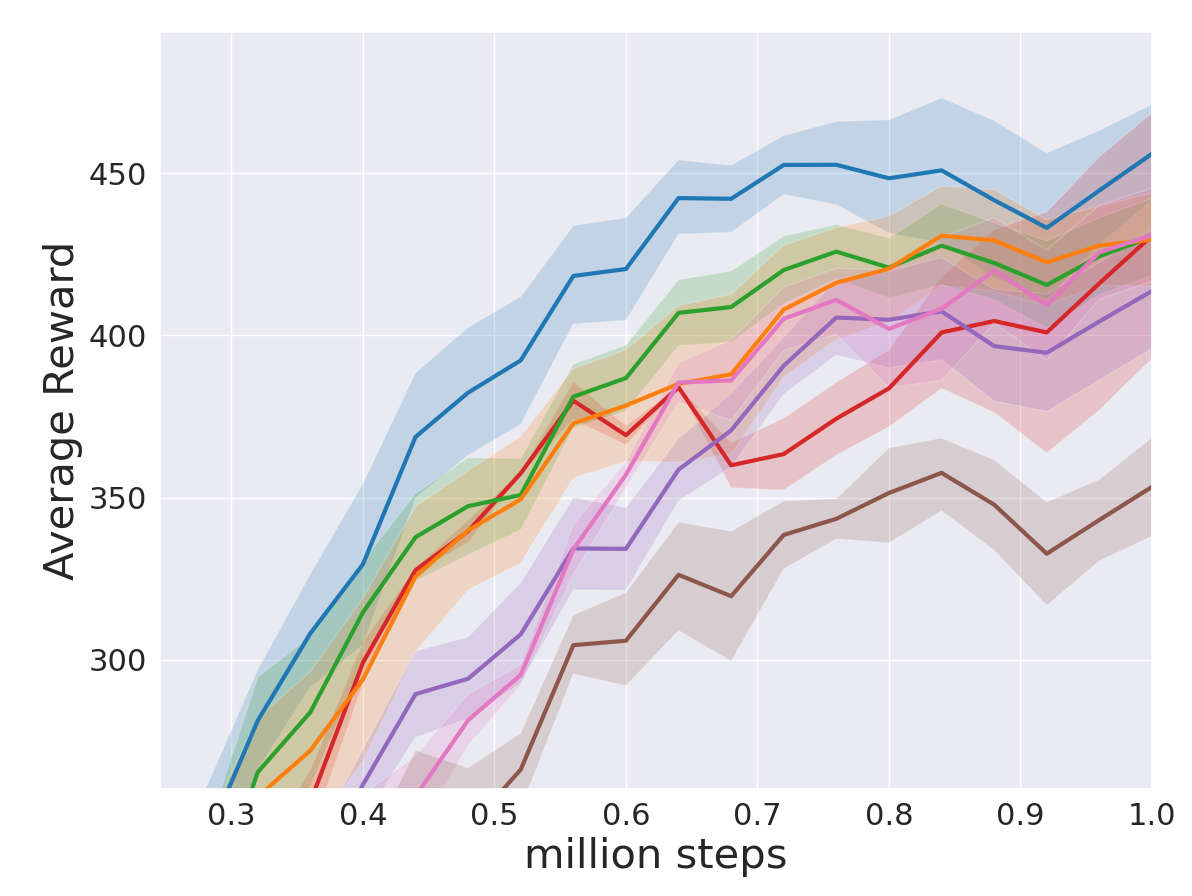}
}
\quad
\subfigure[Pendulum]{
\includegraphics[width=5cm]{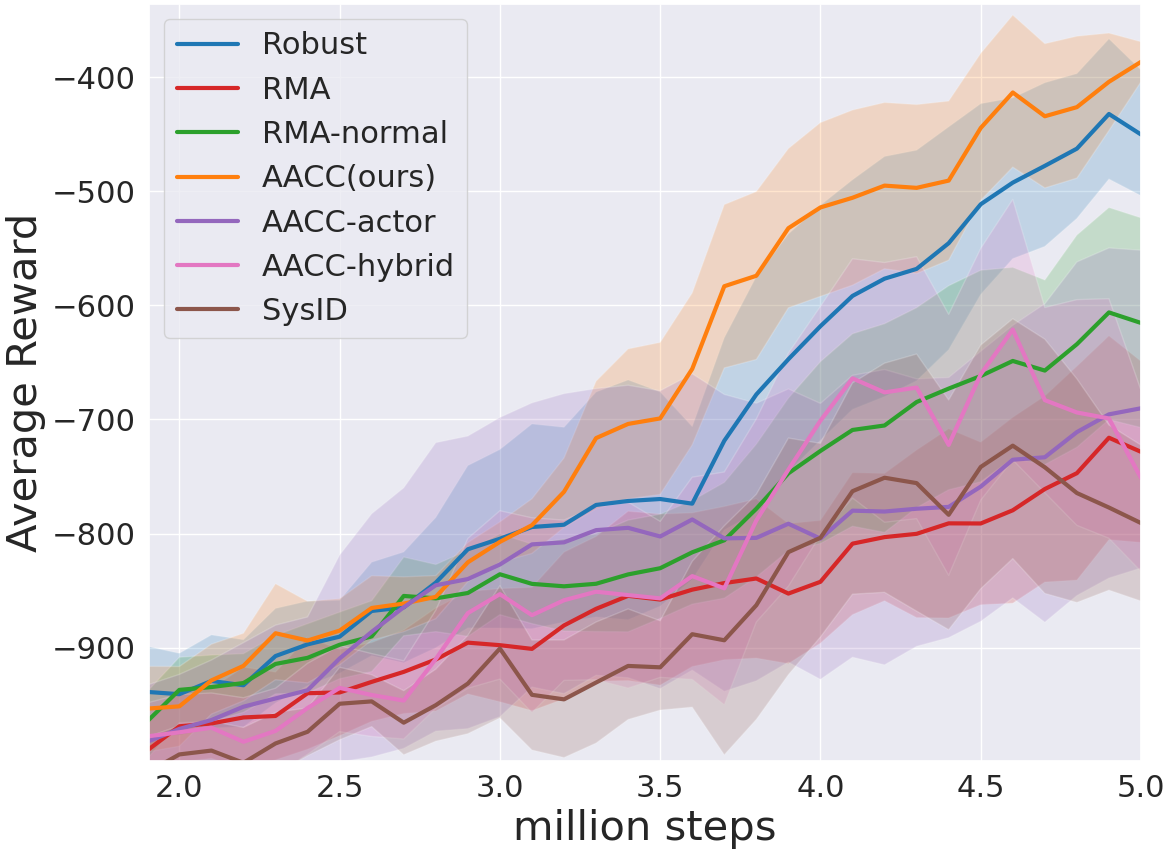}
}
\caption{Evaluation curves every 0.1 million environment steps in easy environments by using different random environmental factors. AACC (yellow) and Robust (blue) performs relatively better than other baselines that the actor takes environmental factors or the context vector as additional inputs.}
\vspace{-10pt}
\label{fig:easy}
\end{figure*}
\begin{figure*}[t]
\centering
\subfigure[Gym-JSBSim]{
\includegraphics[width=5cm]{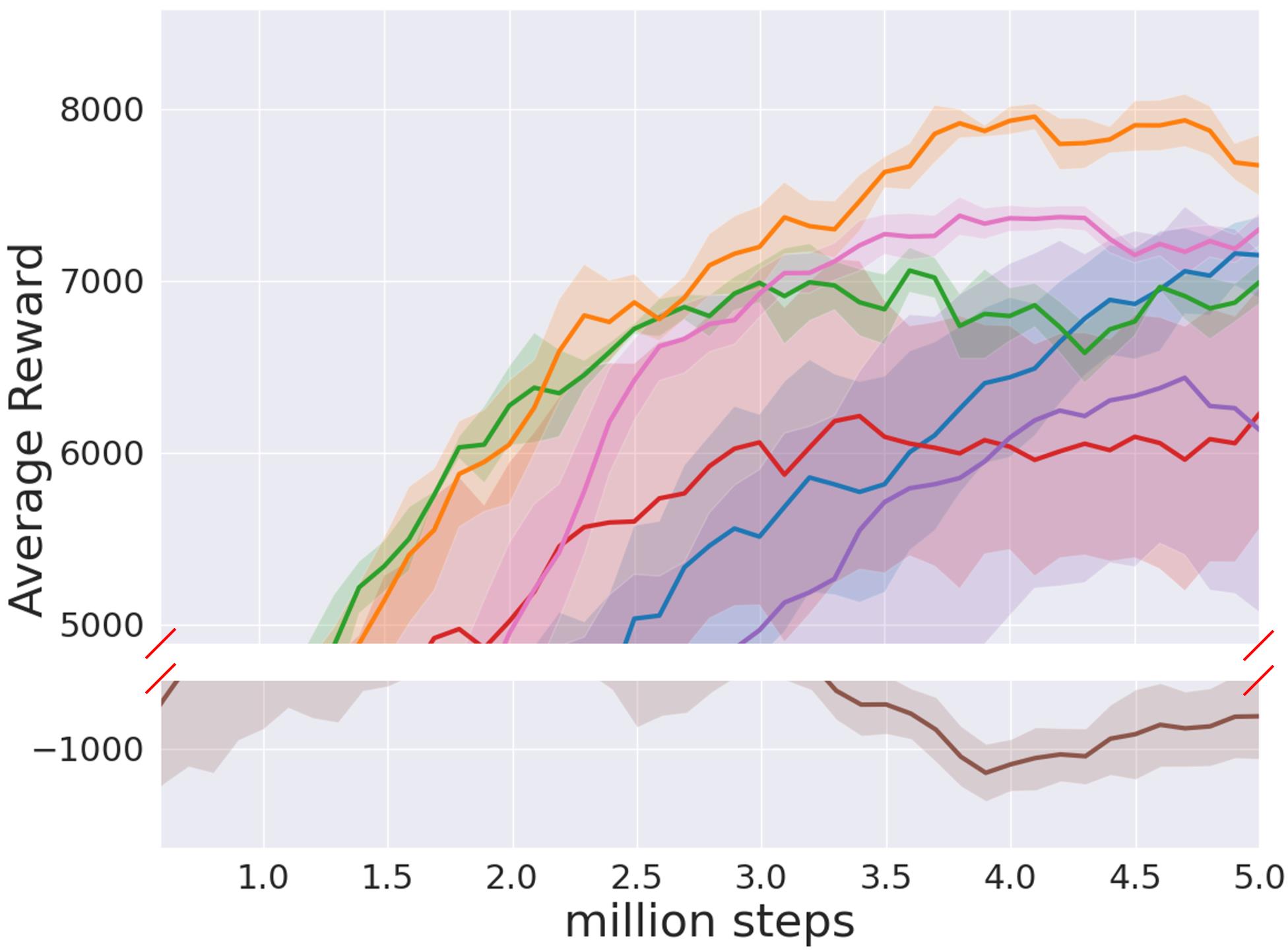}
}
\quad
\subfigure[Hopper]{
\includegraphics[width=5cm]{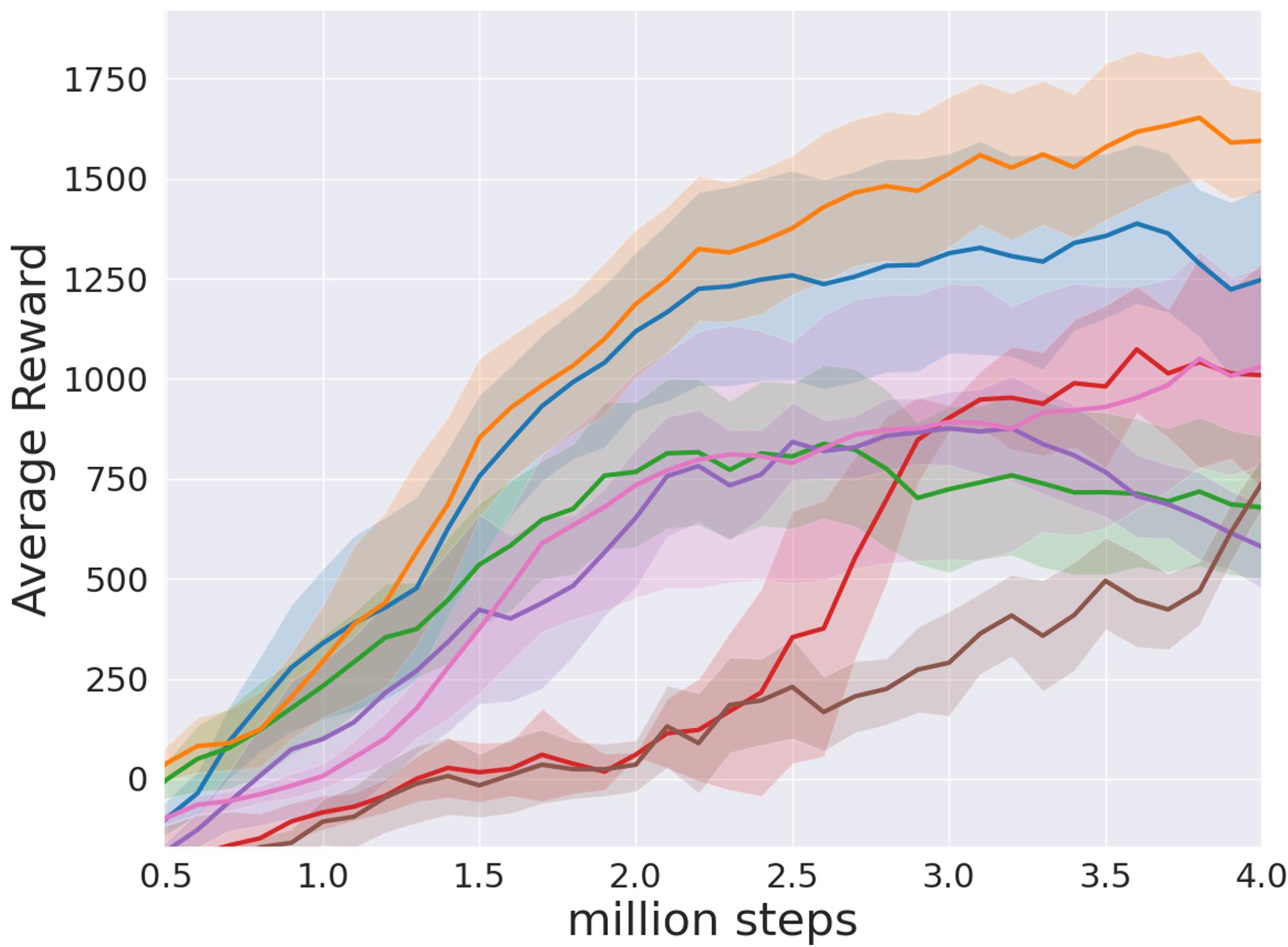}
}
\quad
\subfigure[Laikago]{
\includegraphics[width=5cm]{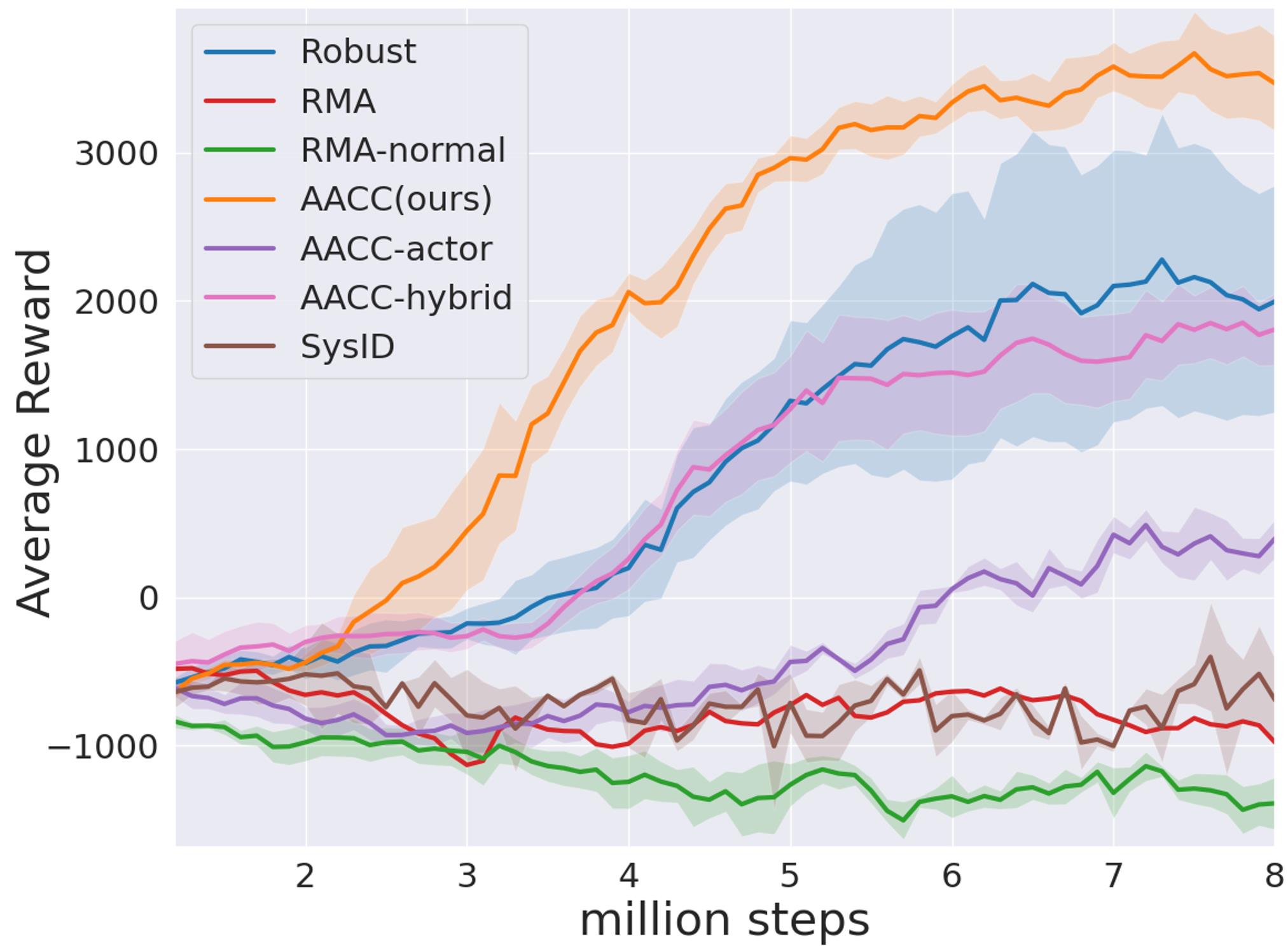}
}
\caption{Evaluation curves every 0.1 million environment steps in middle and hard environments by using different random environmental factors. AACC (yellow) outperforms the baselines in challenging environments. }
\label{fig:hard}
\end{figure*}

\subsection{Comparative Evaluation} \label{Comparative Evaluation}
Figure~\ref{fig:easy} and Figure~\ref{fig:hard} show the total average return of evaluation rollouts during training for AACC and six baselines. We train five different instances of each method with different random environmental factors, with each performing 30 evaluating rollouts every 0.1 million environment steps (in CartPole, every 4 thousand environment steps) by using different random environmental factors. The solid curves correspond to the mean and the standard region to the minimum and maximum returns over the five trials. 

\paragraph{Easy Environments.}
The results are shown in Figure~\ref{fig:easy}. In Acrobot and CartPole, we set the standard deviation of environmental factors values as 2 and 0.5 respectively, which characterizes the severity of the environmental factors change.
We observe that AACC can achieve higher performance in Acrobot that the changes in environmental factors have a high impact on performance. And in CartPole, which is the easiest environment where all methods converge within about 0.7 million time steps, the performance of Robust is better than AACC and other baselines. We attribute it to the bias by the environmental factor encoder that it encodes environmental factors into a context vector. 
We also observe AACC and Robust which do not use environmental factors in the actor network, have better performance than other baselines. It represents that environmental factors used by the actor network (regardless of whether Env Encoder is used or not), do not help the trained policy make a better decision.
To delve deeper into what affects methods on performance, we experimented in harder environments.

\paragraph{Middle and Hard Environments.}
The results are shown in Figure~\ref{fig:hard}.
We observe that the more difficult the environment is, the better performance we obtain with the help from AACC. Whereas, Robust tends to learn a relatively conservative policy due to the lack of environmental factors, which can be totally inferior in both performance and training speed.
Laikago is the hardest environment, in which AACC converges to a better policy and faster than the baseline methods.
Gym-JSBSim is relatively a middle environment, in which methods using both asymmetric critic and Env Encoder can faster convergence than Robust including AACC, AACC-hybrid, and RMA-normal baseline.
Note that, even though RMA-normal and SysID, which are using the symmetric actor-critic method, have extra access to true environmental factors during training and testing, their performance is still limited and even fails to converge, such as in Laikago. It is kind of counter-intuitive that the actor has access to true environmental factors which should perform better based on known environmental factors. We compare with different methods described above for further analyzing the reason. The result shows that introducing the environmental factors only has a positive effect on the critic part, while doing the same can occur a negative performance for the actor part (AACC vs. AACC-actor, AACC-hybrid, RMA-normal; Robust vs. SysID),  in addition, employing the environmental factor encoder can improve the performance of the method (RMA-normal vs. SysID). Finally, we compare RMA-normal with RMA, which is proposed in \citet{kumar2021rma} to deal with quadruped robots that can be zero-shot transferred to real world scenarios. However, in our environment settings, RMA-normal has a better performance than RMA.

The results show that, overall, AACC performs comparably to the baselines on the easier tasks and outperforms them on the harder tasks with a large margin, both in terms of learning speed and the final performance.

\begin{figure*}[tbp]
\centering
\subfigure[Encoding Dimension]{
\includegraphics[width=5cm]{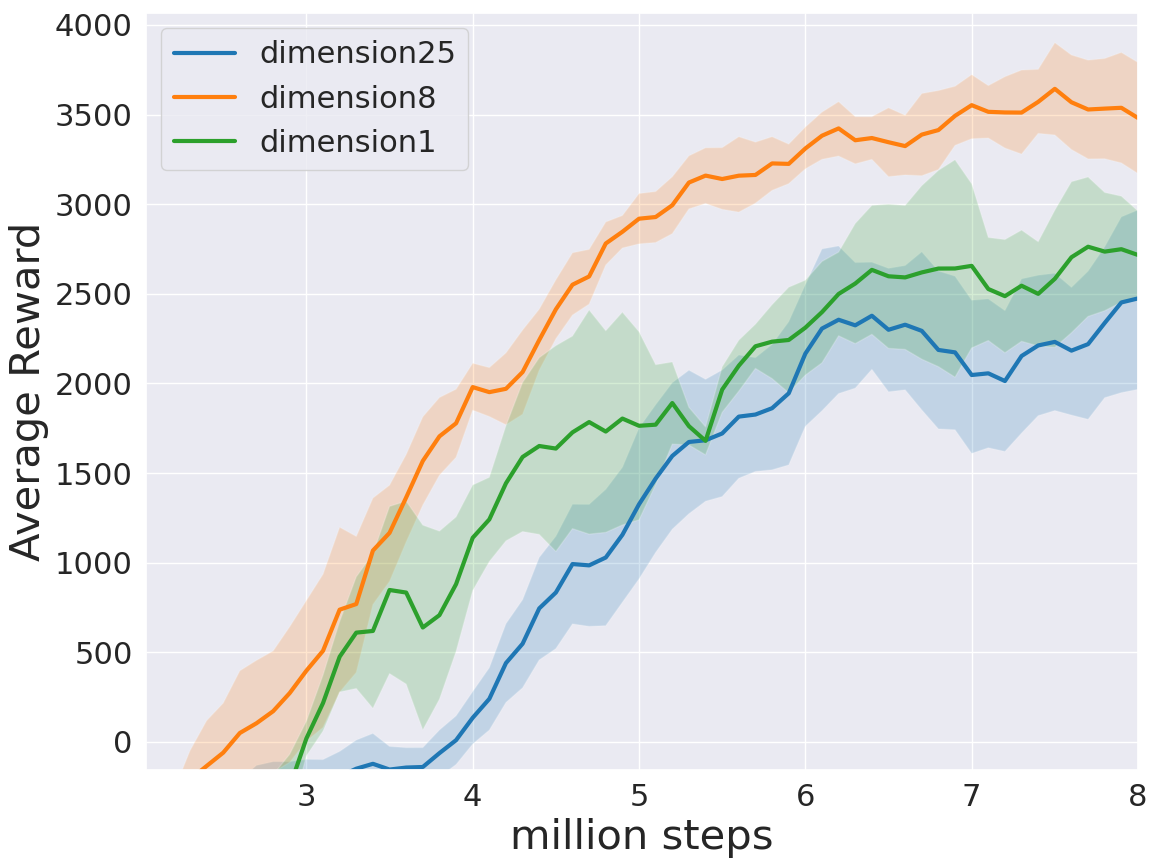}
\label{encoding dimension}
}
\quad
\subfigure[Env Encoder]{
\includegraphics[width=5cm]{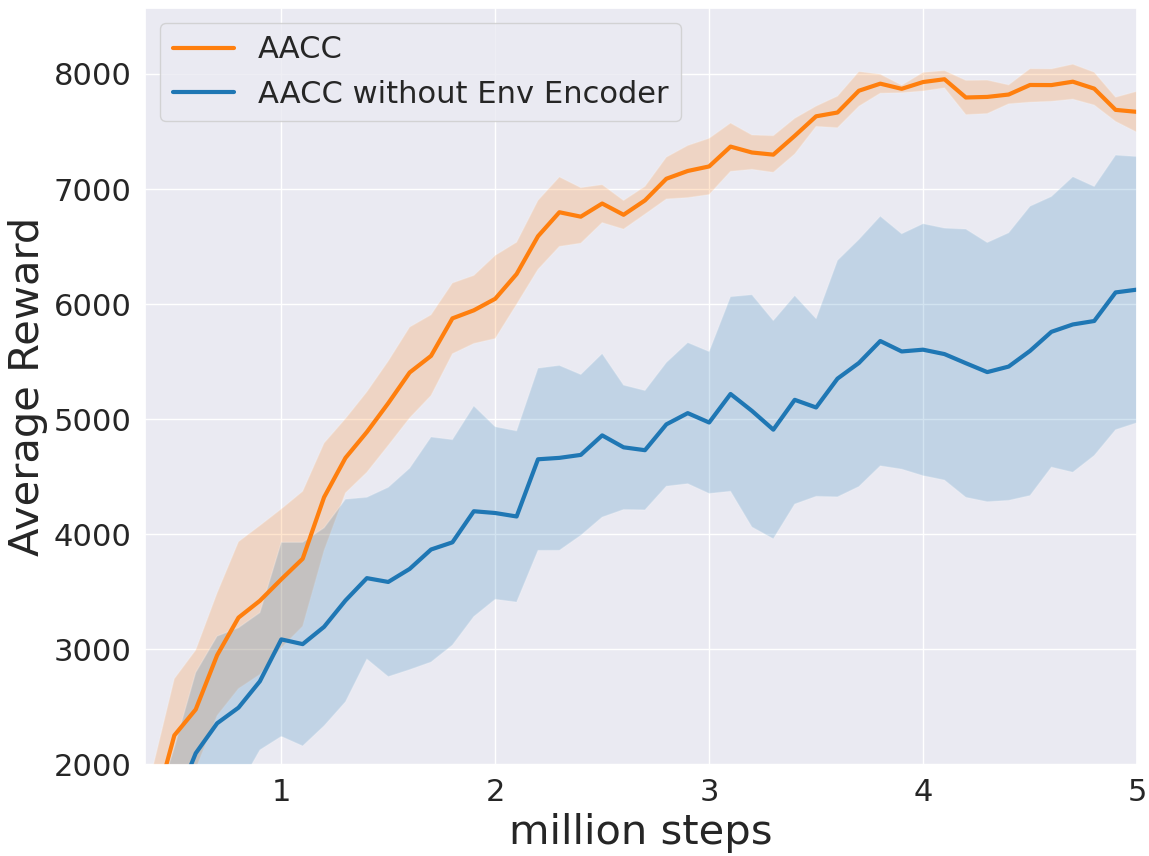}
\label{encoder}
}
\quad
\subfigure
[Domain Randomization]{
\includegraphics[width=5cm]{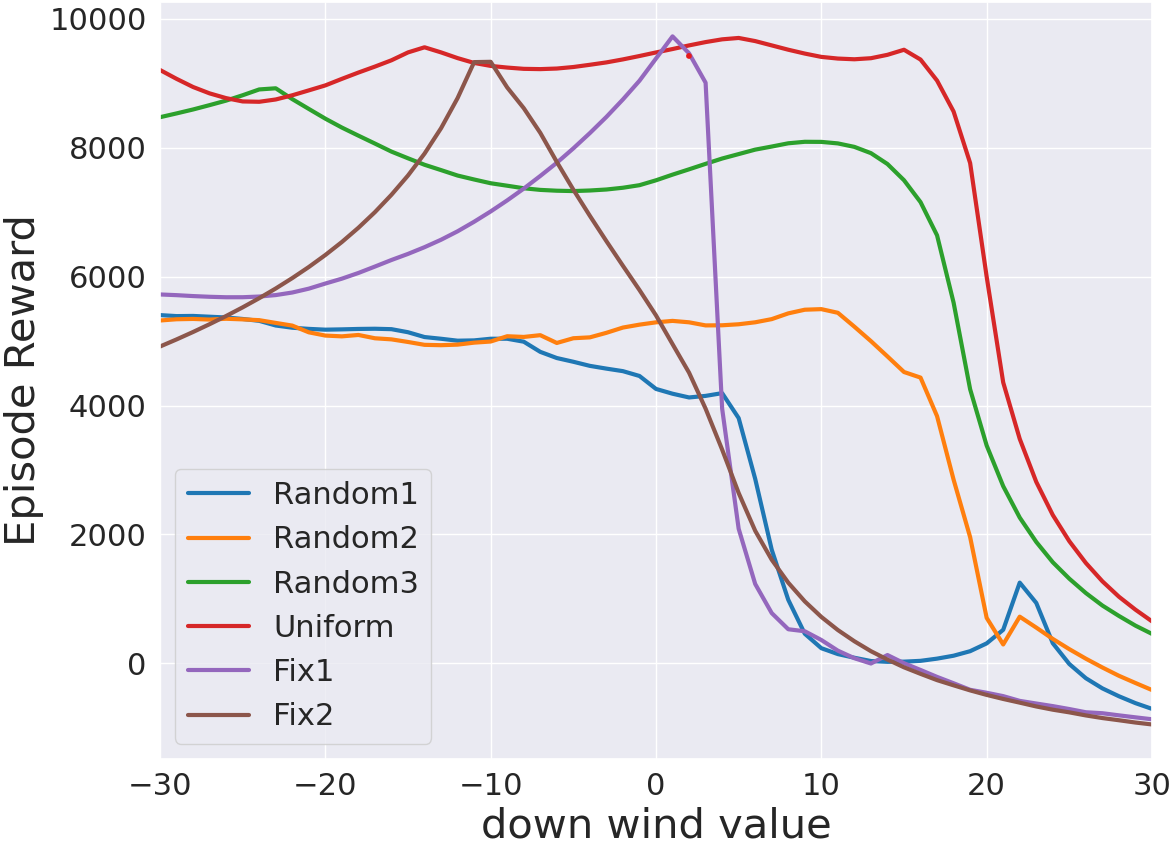}
\label{random}
}
\caption{Figure~(a) shows an evaluating curve of AACC with different encoding dimension in Laikago. Figure~(b)  shows evaluating curve of AACC with or without Env Encoder that environmental factor encoder can improve performance. Figure~(c) shows how domain randomization affects the generalization of the method in the test environments by using some different settings of the change of the environmental factors.}
\label{fig: ablation result}
\vspace{-10pt}
\end{figure*}

\subsection{Ablation Study}
In this section, we further examine which particular component of AACC is important for a good performance. We also examine how sensitive AACC is to hyperparameters, namely the encoding dimension of environmental factors. Finally, we use some different change settings of environmental factors in training to evaluate how domain randomization affects generalization in test environments.
\vspace{-10pt}
\paragraph{Encoding Dimension.}
To assess the impact of different dimension of the output of Env Encoder on performance, we experiment in Laikago, which has 17 experiment parameters, making encoding dimensions more widely available. In Figure~\ref{encoding dimension}, we compare the performance of AACC when the dimension of the context vector $c$ varies. We find that it has the highest performance when the dimension of $c$ is about half the dimension of the environmental factors. The dimension of $c$ is too large or too small to be desirable. 
When the dimension of $c$ is 1, the learning curve is unstable, and when it is 25, the training speed is limited.
This result shows that encoding the environmental factors to a lower dimension can effectively avoid redundant information interfering with the convergence of the critic network, since some of the environmental factors have the same, or at least similar effect on the training process, while some others do not have any effect at all. 
To further evaluate the effect of each environmental factor on the training process, we pick up one certain environmental factor as the input. The details are shown in Appendix~\ref{D1}.

\paragraph{With vs. Without Env Encoder.} \label{env encoder}
Comparative evaluation has shown the role of the environmental factor encoder by comparing the RMA-normal and SysID baseline in the case of multiple environmental factors. For example, SysID fails to make any progress in Gym-JSBSim, while RMA-normal converges to the sub-optimal policy. To evaluate the role of the environmental factor encoder in AACC, we experiment in Gym-JSBSim, Hopper and Laikago environments where the dimension of environmental factors is 3, 11, 17, respectively. Figure~\ref{encoder} and Figure~\ref{without encoder} (Appendix~\ref{additional experiment}) compare AACC to AACC without environmental factor encoder that inputs environmental factors $e$ directly into the state $s_t=(o_t, e)$. It indicates that adding an environmental factor encoder into asymmetric critic can yield better performance and accelerate learning.
\vspace{-8pt}
\paragraph{Domain Randomization.}
To evaluate how domain randomization affects generalization in test environments, we use some different change settings of environmental factors \emph{i.e.}, change down wind value between episodes in Gym-JSBSim.
Random1, Random2 and Random3 represent AACC is trained in down wind value varying in -30 or 30, and -30, 0, or 30, and -30, -15,...,30, respectively. 
Uniform means down wind value is sampled in the range of -30 to 30, Fix1 and Fix2 are respectively fixed the down wind value in -10 and 0. In the testing phase, the down wind value is sampled uniformly in the range of -30 to 30 and other environmental factors do not change. 
The result is shown in Figure~\ref{random}.
We observe that Fix1 and Fix2 can not generalize to a test environment that differs from the training environment and Uniform has a better performance than Random1, Random2 and Random3.
It shows that sufficient domain randomization is necessary. 
Meanwhile, we observe that the reward of Uniform is not worse than Fix1 and Fix2. This result shows that domain randomization does not make AACC performance conservative.

\begin{table*}[tbp]
    \centering
    \caption{Mean evaluation rewards $\pm$ one standard derivation in Gym-JSBSim and Laikago. Different Distribution means the environmental factors during evaluation follow new distributions. Unseen scenarios mean that the environmental factors are slightly shifted during evaluation.}
    
    \begin{tabular}{l|lllllll}
    \toprule
        \multirow{2}{*}{} & \multicolumn{7}{c}{Different Distribution}                                                      \\ \cline{2-8} 
                   & AACC               & Robust    & RMA          & RMA-normal & AACC-actor  & AACC-hybrid & SysID         \\ \hline
        Gym-JSBSim & \textbf{8485±404}  & 7688±713  & 7077±860     & 6652±2656  & 6996±1176   & 7879±520    & -1059±676     \\
        Laikago    & \textbf{3271±715}  & 2380±912  & -611±128     & -1031±291  & 956±628     & 2065±543    & -980±400      \\ \hline
        \multirow{2}{*}{} & \multicolumn{7}{c}{Unseen Scenarios}                                                            \\ \hline
        Gym-JSBSim & \textbf{8353±623}  & 8017±1147 & 7059±1200    & 5270±2603  & 4139±1747   & 7355±1020   & -2276±1361    \\
        Laikago    & \textbf{1789±1047} & 1492±871  & -572±63      & -947±188   & -613±204    & -319±273    & -942±488      \\
    \bottomrule
    \end{tabular}
    \label{new scenarios table}
\end{table*}

\begin{table*}[tbp]
    \centering
    \caption{Mean evaluation rewards $\pm$ one standard derivation of different AACC-hybrid in Laikago. hybrid(x) indicates that the output
dimension of Env Encoder to the actor is x.}
    
    \begin{tabular}{c|lllllll}
    \toprule
                & AACC & hybrid(1) & hybrid(4) & hybrid(8) & hybrid(17) & hybrid(25) \\ \hline
     Laikago    & 3530±680 & 2607±757 & 1826±687 & 1891±384 & 1430±680 & 1353±474 \\
    \bottomrule
    \end{tabular}
    \label{dimension of actor table}
\end{table*}

\subsection{Scenarios of training and test are different} \label{sec: Different scenarios}
To further verify that our method can make the agent generalize to scenarios different from the training set, we conduct two different sets of experiments. For the case where the environmental factors during evaluation follow new distributions, we put truncated normal distributions on all environmental factors within their original bounds during testing. As for unseen scenarios, we take the slightly shifted uniform distributions for environmental factors during evaluation. The details of experimental setting can be found in Table~\ref{new scenarios} (Appendix~\ref{appe:environment description}).
The results are shown in Table~\ref{new scenarios table}, Figure~\ref{Different Distribution} and Figure~\ref{Unseen Scenarios} (Appendix~\ref{additional experiment}). We observe that AACC outperforms all the baselines in both two cases, although these two cases are more difficult than the case in Section~\ref{Comparative Evaluation}, where training and test environment share the same distribution of environmental factors. Note that AACC-actor and AACC-hybrid have a positive evaluation reward in the case of different distribution in Laikago, but have a negative evaluation reward. This result shows that AACC-actor and AACC-hybrid do not learn the ability to generalize to unseen scenarios in Laikago.

\subsection{Impact of environmental factors on the actor}
To further demonstrate the counter-intuitive phenomenon that policies that have access to true environmental factors may not perform better than policies that do not make decisions based on environmental factors, we experiment with varying the dimension of the output of Env Encoder to the actor, with the one to the critic remaining unchanged. There are 17 environmental factors in Laikago, and hybrid(x) in Table~\ref{dimension of actor table} indicates that the output dimension of Env Encoder to the actor of AACC-hybrid is x. In particular, hybrid(0) is exactly AACC. Under the same training and test setup, the results are shown in Table~\ref{dimension of actor table} and Figure~\ref{dimension of actor} (Appendix~\ref{additional experiment}). Surprisingly, the larger the output dimension of Env Encoder to the actor network is, the worse the performance is. The experimental result shows that the information of environmental factors added to the actor network cannot help the agent to take suitable actions, or may even impose a negative impact. 

\paragraph{Discussion.} The environmental factors are different from the observation in an MDP. The value of the environmental factors does not change depending on the action, but only on the user's settings for the environment (such as randomly sampling from distribution at the beginning of one round). Therefore, we think that environmental information and observation may not be simply treated equally, such as taken as input to the actor network. Moreover, the training of the critic network is a supervised learning process, and its objective function is to minimize the state value error, so the critic network with environmental information can more accurately assess the state value. However, the actor network is updated by the policy gradient, based on the comparison of the previous point, we guess that there may be a problem simply giving environmental factors and observation as state information to the actor network.

\section{Conclusion}
We propose AACC in Contextual RL, an end-to-end actor-critic RL method in which the critic is trained with environmental factors and observation while the actor only gets the observation as input. 
The theoretical results enable convergence in the case of the asymmetric critic. 
Moreover, we experimentally show that such a method enables the policy to converge to a better policy faster than the baselines. Finally, we show which component of the method affect the performance experimentally, and how environmental factors affect the actor and the critic respectively.

\bibliography{uai2022-template}

\newpage
\appendix
\providecommand{\upGamma}{\Gamma}
\providecommand{\uppi}{\pi}

\newpage
\begin{onecolumn}

\section{Proof}\label{appe:proof}

\textbf{Theorem}~\ref{theorem1}.
For any given CMDP and policy $\pi$, $V_{\pi}(c,o)$ is an unbiased estimation of $V_{\pi}(o)$, i.e., $V_{\pi}(o)=\mathbb{E}_{c \sim p(c)}[V_{\pi}(c,o)]$, assuming that the policy is determined by the observation $o$, i.e., $\pi(c,o)=\pi(o)$.
\begin{proof}
The actor $\pi$ only get observation and it takes action only depend on observation, so $\pi(a|o)=\pi(a|c,o)$

\begin{equation}
\begin{aligned}
\mathbb{E}_{c \sim p(c)}\left[V_{\pi}(c,o)\right]&= \mathbb{E}_{c \sim p(c)}\left[\sum_a \pi(a|c,o)Q_{\pi}(c,o,a)\right]\\
&= \mathbb{E}_{c \sim p(c)}\left[\sum_a \pi(a|o)Q_{\pi}(c,o,a)\right] \\
&= \sum_a \pi(a|o)\mathbb{E}_{c \sim p(c)}\left[Q_{\pi}(c,o,a)\right] \\
&= \sum_a \pi(a|o)\mathbb{E}_{c \sim p(c)}\left[R(c,o,a)+\gamma \mathbb{E}_{o'|c, o, a} \left[V_{\pi}(c,o') \right]\right] \\
&= \sum_a \pi(a|o)(R(o,a) + \gamma \mathbb{E}_{c \sim p(c)}\left[\mathbb{E}_{o'|c, o, a} \left[V_{\pi}(c,o')\right]\right] \\
&= \sum_a \pi(a|o)(R(o,a) + \gamma \mathbb{E}_{o'|o, a}\left[V_{\pi}(o')\right]) \\
&= \sum_a \pi(a|o) Q_{\pi}(o, a) \\
&= V_{\pi}(o)
\end{aligned}
\end{equation}
\end{proof}

\textbf{Theorem}~\ref{theorem2}.(Policy Gradient in AACC).
Assume that the context $c$ follows a distribution $p(c)$ at the beginning of an episode, and remains fixed within the same episode, then we have:
\begin{equation}
    \nabla J(\pi) = \mathbb{E}_{c \sim p(c)}\left[\mathbb{E}[Q_{\pi}(c,o,a)\nabla \log\pi(a|o)]\right].
\end{equation}

\begin{proof}
 Following Theorem~\ref{theorem1}, we have $Q_{\pi}(o,a)= \mathbb{E}_{c \sim p(c)}[Q_{\pi}(c,o,a)]$.

Therefore,
\begin{equation}
\begin{aligned}
\nabla J(\pi) &= \nabla \mathbb{E}_{c \sim p(c)}\left[\sum_a\pi(a|o)Q_{\pi}(c, o, a)\right] \\
&= \nabla\left[\sum_a\pi(a|o)Q_{\pi}(o, a)\right] \\
&= \mathbb{E}[Q_{\pi}(o,a)\nabla\log\pi(o,a)] \quad\quad\quad\quad \text{see Equation~\ref{gradient}.} \\
&= \mathbb{E}_{c \sim p(c)}\left[\mathbb{E}\left[Q_{\pi}(c,o,a)\nabla\log\pi(a|o)\right]\right] \\
\end{aligned}
\end{equation}
\end{proof}

\newpage

\section{Environment Descriptions}\label{appe:environment description}
Our experiments include six environments, \emph{i.e.}, Gym-JSBSim, Laikago, Hopper, CartPole, Acrobot and Pendulum. Gym-JSBSim provides RL environments for the control of fixed-wing aircraft using the JSBSim \citep{2004JSBSim} flight dynamics model. In Laikago, the agent is a 18DoF Unitree Laikago quadruped. In Hopper, the agent is a 4-link, 6DoF 2D hopper. CartPole, Acrobot and Pendulum are contextually extended classic control from OpenAI Gym \citep{2016OpenAI}. See Figure~\ref{illustration} for an overview of included environments. Table~\ref{environment list} lists relative information of these environments.

\begin{figure*}[ht]
\centering
\subfigure[Acrobot]{
\includegraphics[width=5cm]{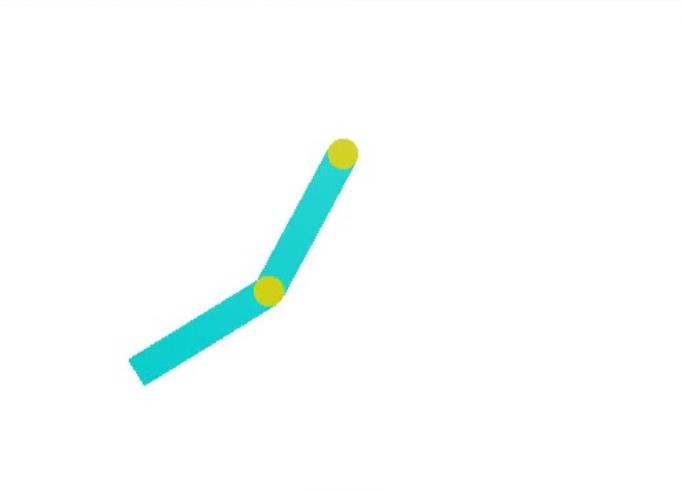}
}
\quad
\subfigure[CartPole]{
\includegraphics[width=5cm]{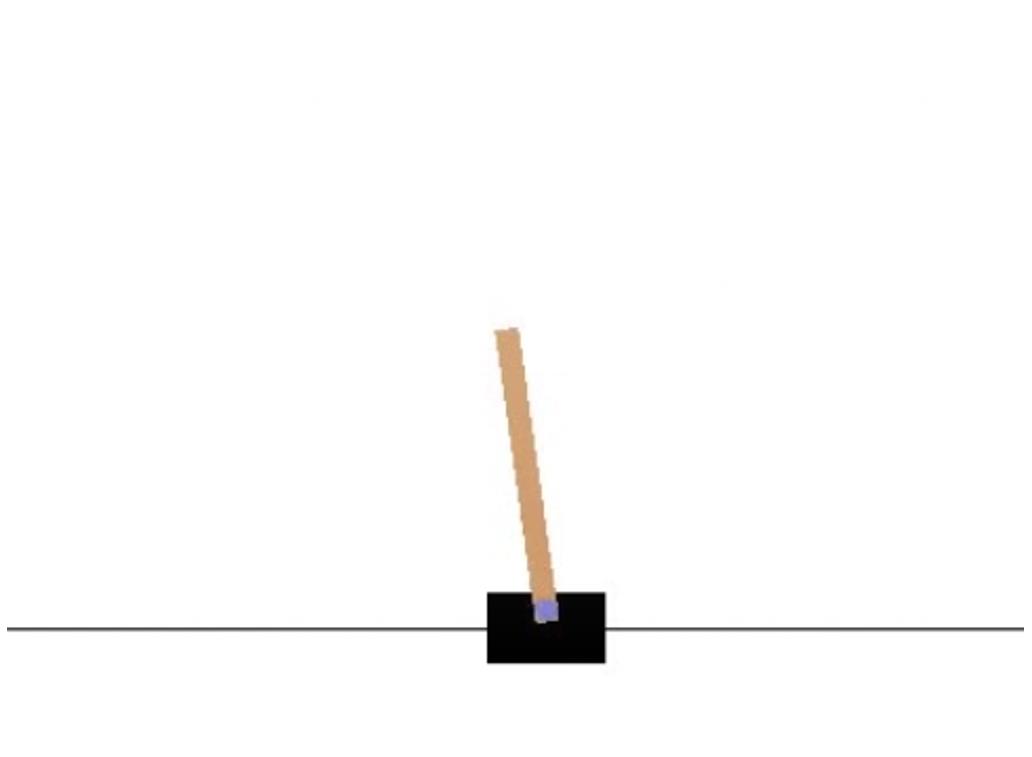}
}
\quad
\subfigure[Pendulum]{
\includegraphics[width=5cm]{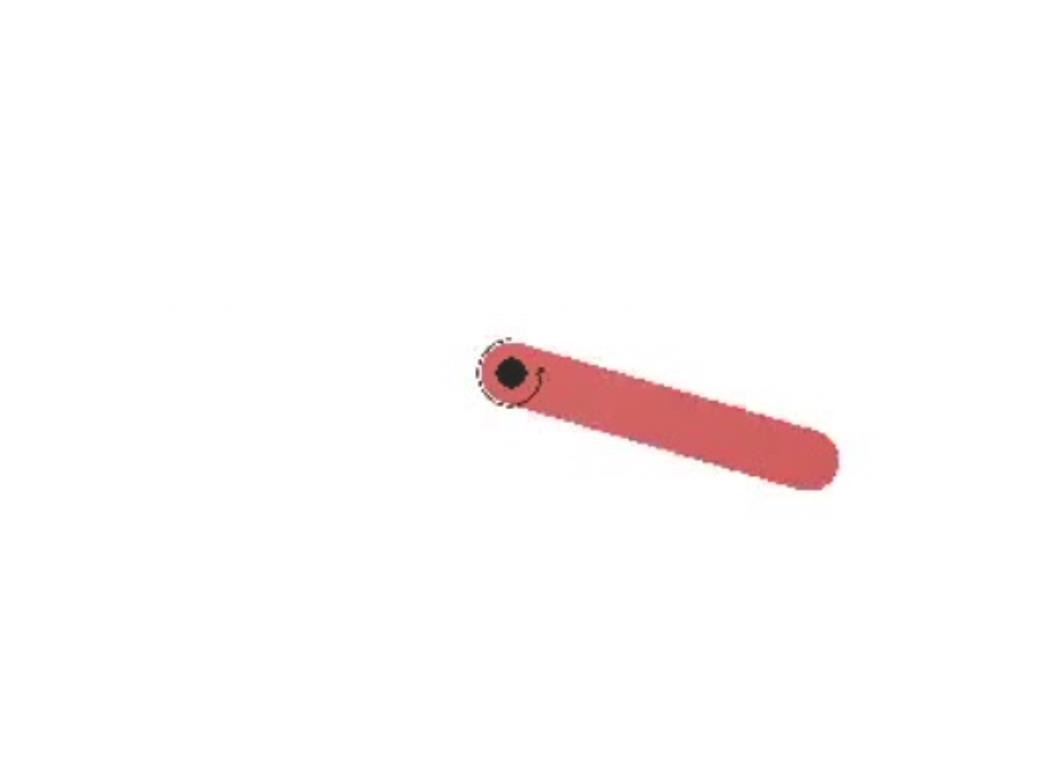}
}

\subfigure[Gym-JSBSim]{
\includegraphics[width=5cm]{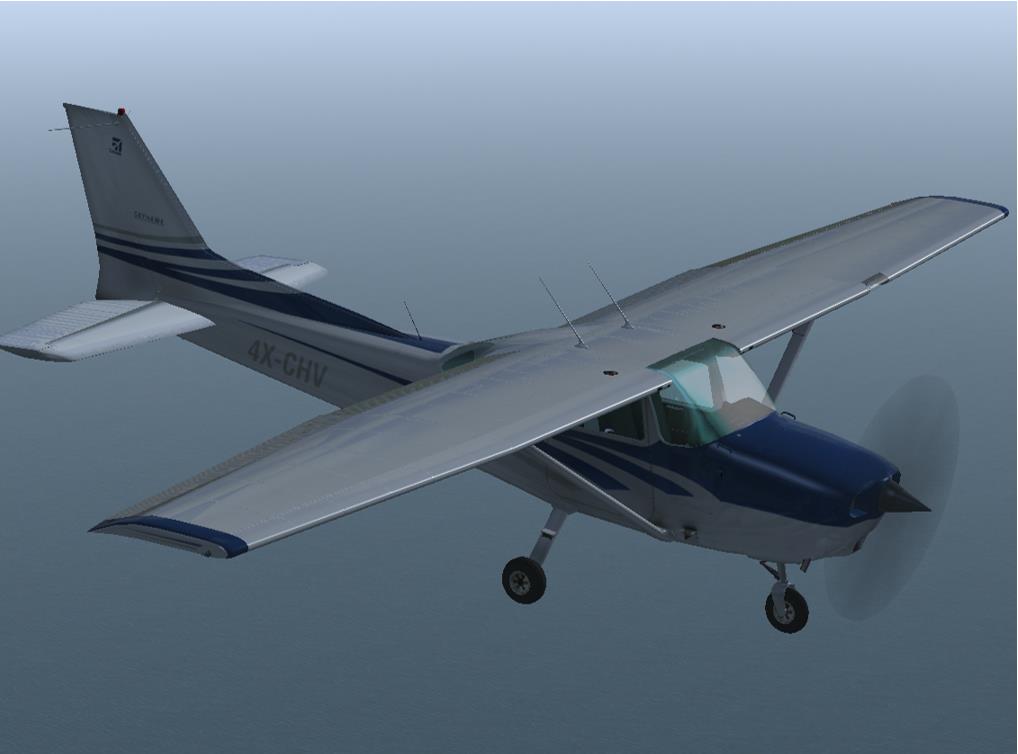}
}
\quad
\subfigure[Hopper]{
\includegraphics[width=5cm]{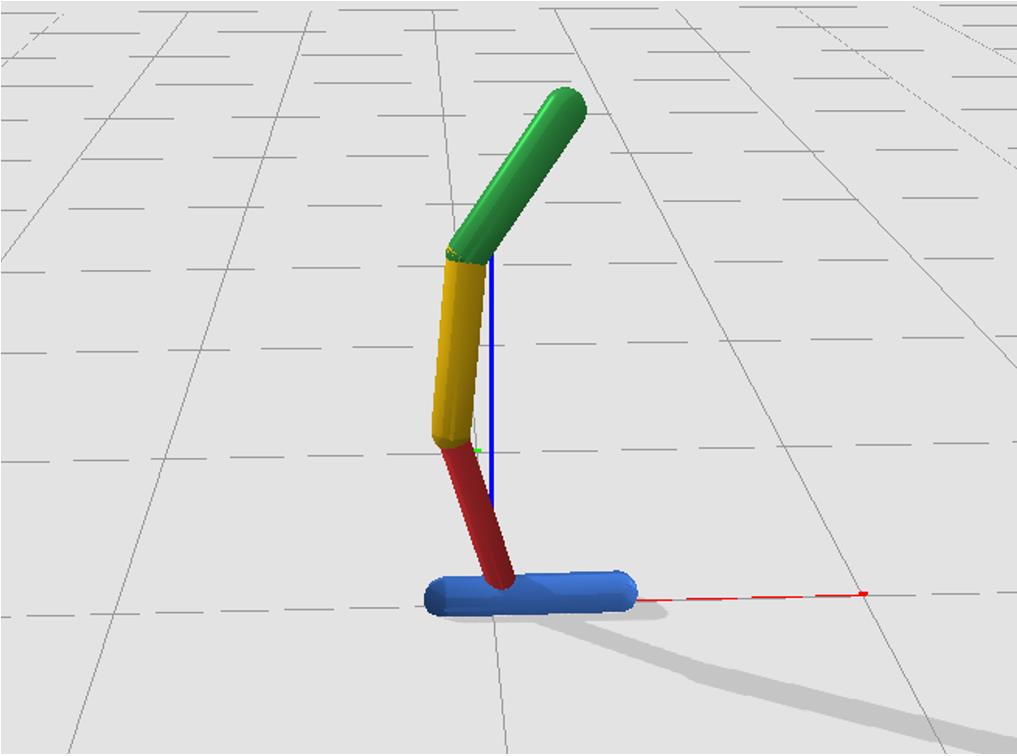}
}
\quad
\subfigure[Laikago]{
\includegraphics[width=5cm]{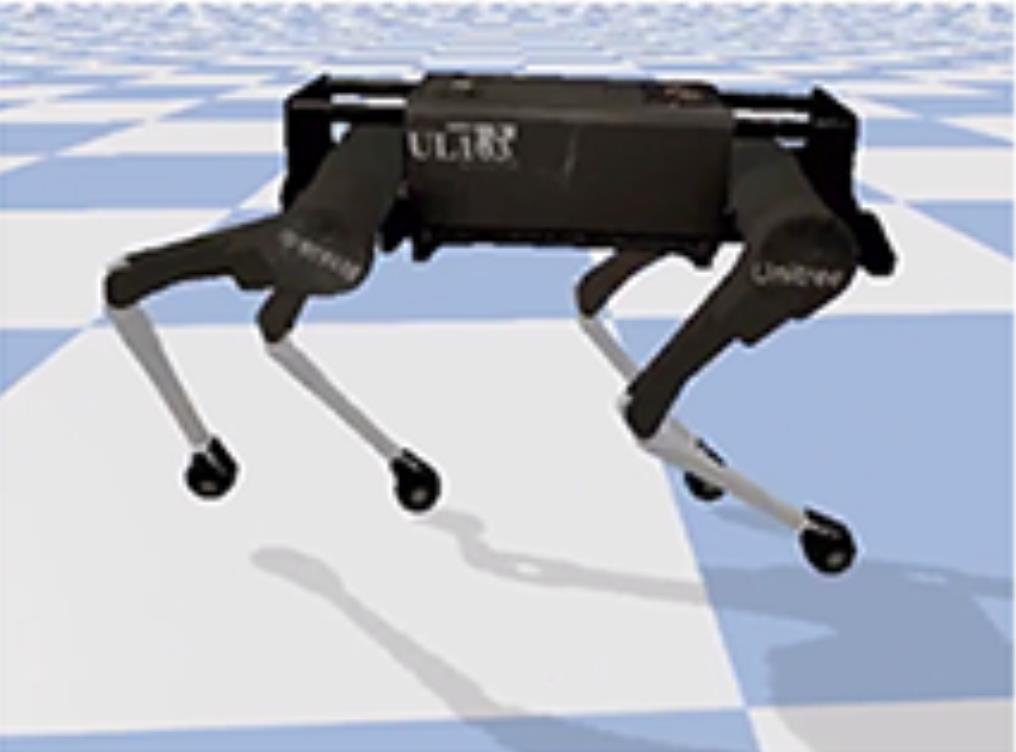}
}
\caption{Illustrations of six environments.}
\label{illustration}
\end{figure*}

\begin{table}[ht]
    \centering
    \caption{\textbf{Environment List.} \textbf{C/D} stands for continuous action or discrete action. \textbf{Dimension} stands for the dimension of the environmental factors. \textbf{Difficulty}  stands for the difficulty of a specific environment task.}
    \begin{tabular}{lccccll}
      \toprule
      \bfseries Name & \bfseries State Space & \bfseries Action Space & \bfseries C/D & \bfseries Dimension & \bfseries Difficulty & \bfseries Platform \\
      \midrule
      Acrobot & 6 & 3 & D & 7 & easy & Gym\\
      CartPole & 4 & 2 & D & 6 & easy & Gym\\
      Pendulum & 3 & 1 & C & 4 & easy & Gym\\
      Gym-JSBSim & 16 & 4 & C & 3 & middle & JSBSim\\
      Hopper & 11 & 3 & C & 11 & middle & Pybullet\\
      Laikago & 46 & 12 & C & 17 & hard & Pybullet\\
      \bottomrule
    \end{tabular}
    \label{environment list}
\end{table}

\vspace{2in}

\begin{table}[p]
    \centering
    \caption{\textbf{Default} stands for default value. \textbf{Std} stands for the standard deviation.}

\subtable[Acrobot]{
    \begin{tabular}{lrllll}
    \toprule
    \bfseries Name & \bfseries Default & \bfseries Bounds & \bfseries Type & \bfseries Distribution & \bfseries Std \\
    \midrule
    link\_com\_1 & 0.50 & (0, 1) & \multirow{7}{*}{float} & \multirow{7}{*}{Gaussian} & \multirow{7}{*}{2}\\
    link\_com\_2 & 0.50 & (0, 1)\\
    link\_length\_1 & 1.00 & (0.1, 10)\\
    link\_length\_2 & 1.00 & (0.1, 10)\\
    link\_mass\_1 & 1.00 & (0.1, 10)\\
    link\_mass\_2 & 1.00 & (0.1, 10)\\
    link\_moi & 1.00 & (0.1, 10)\\
    \bottomrule
    \end{tabular}}
    
    \subtable[CartPole]{
    \begin{tabular}{lrllll}
      \toprule
      \bfseries Name & \bfseries Default & \bfseries Bounds & \bfseries Type & \bfseries Distribution & \bfseries Std \\
      \midrule
      force\_magnifier & 10.00 & (1, 100) & int & \multirow{6}{*}{Gaussian} & \multirow{6}{*}{0.5}\\
      gravity & 9.80 & (0.1, inf) & float\\
      masscart & 1.00 & (0.1, 10) & float\\
      masspole & 0.10 & (0.01, 1) & float\\
      pole\_length & 0.50 & (0.05, 5) & float\\
      update\_interval & 0.02 & (0.002, 0.2) & float\\
      \bottomrule
    \end{tabular}
    }
    
    \subtable[Pendulum]{
    \begin{tabular}{lrllll}
      \toprule
      \bfseries Name & \bfseries Default & \bfseries Bounds & \bfseries Type & \bfseries Distribution & \bfseries Std \\
      \midrule
      dt & 0.05 & (0, inf) & \multirow{4}{*}{float} & \multirow{4}{*}{Gaussian} & \multirow{4}{*}{0.1}\\
      g & 10.00 & (0, inf) \\
      l & 1.00 & (1e-06, inf)\\
      m & 1.00 & (1e-06, inf)\\
      \bottomrule
    \end{tabular}
    }
    
    \subtable[Gym-JSBSim]{
    \begin{tabular}{lcrll}
      \toprule
      \bfseries Name & \bfseries Dimension & \bfseries Bounds & \bfseries Type & \bfseries Distribution \\
      \midrule
      north\_wind & 1 & (-30, 30) & \multirow{3}{*}{float} & \multirow{3}{*}{Uniform}\\
      east\_wind & 1 & (-30, 30) \\
      down\_wind & 1 & (-30, 30) \\
      \bottomrule
    \end{tabular}
    }
    
    \subtable[Hopper]{
    \begin{tabular}{lcrll}
      \toprule
      \bfseries Name & \bfseries Dimension & \bfseries Bounds & \bfseries Type & \bfseries Distribution \\
      \midrule
      mass ratio & 4 & (1, 3) & \multirow{5}{*}{float} & \multirow{5}{*}{Uniform}\\
      inertia ratio & 4 & (0.3, 1.8) \\
      lateral friction & 1 & (0.5, 5) \\
      restitution & 1 & (0, 0.5) \\
      spinning friction & 1 & (0, 0.2) \\
      \bottomrule
    \end{tabular}
    }
    
    \subtable[Laikago]{
    \begin{tabular}{lcrll}
      \toprule
      \bfseries Name & \bfseries Dimension & \bfseries Bounds & \bfseries Type & \bfseries Distribution \\
      \midrule
      mass ratio & 1 & (1, 2) & \multirow{5}{*}{float} & \multirow{5}{*}{Uniform}\\
      inertia ratio & 13 & (0.5, 1.5) \\
      lateral friction & 1 & (0.1, 5) \\
      restitution & 1 & (0, 0.5) \\
      spinning friction & 1 & (0, 0.2) \\
      \bottomrule
    \end{tabular}
    }
\end{table}

\begin{table}[ht]
    \centering
    \caption{The experimental setups of Section~\ref{sec: Different scenarios}: the distribution of each environmental factor during training and testing.}
    \subtable[Gym-JSBSim]{
    \begin{tabular}{lccc}
      \toprule
      \bfseries \multirow{2}{*}{Name} & \bfseries \multirow{2}{*}{Train} & \multicolumn{2}{c}{ \bfseries Test} \\ \cline{3-4}
      \bfseries & \bfseries & \bfseries New Distribution & \bfseries Unseen Scenarios \\
      \midrule
      north\_wind & \multirow{3}{*}{Uniform(-30, 30)} & \multirow{3}{*}{Normal(0, 15)} & \multirow{3}{*}{Uniform(-35, -30)} \\
      east\_wind \\
      down\_wind \\
      \bottomrule
    \end{tabular}
    }
    
    \subtable[Laikago]{
    \begin{tabular}{llll}
      \toprule
      \bfseries \multirow{2}{*}{Name} & \bfseries \multirow{2}{*}{Train} & \multicolumn{2}{c}{ \bfseries Test} \\ \cline{3-4}
      \bfseries & \bfseries & \bfseries New Distribution & \bfseries Unseen Scenarios \\
      \midrule
      mass ratio & Uniform(1, 2) & Normal(1.5, 0.5) & Uniform(2, 2.2) \\
      inertia ratio (13dim) & Uniform(0.5, 1.5) & Normal(1, 0.25) & Uniform(1.5, 1.8) \\
      lateral friction & Uniform(0.1, 5) & Normal(2.5, 1.5) & Uniform(5, 6) \\
      restitution & Uniform(0, 0.5) & Normal(0.25, 0.15) & Uniform(0.5, 0.7) \\
      spinning friction & Uniform(0, 0.2) & Normal(0.1, 0.05) & Uniform(0.2, 0.25) \\
      \bottomrule
    \end{tabular}
    }
    \label{new scenarios}
\end{table}

\newpage

\section{Variant Method}\label{variant}
Figure~\ref{aacc-actor} shows that Inverse Asymmetric Actor-Critic (AACC-actor) adds Env Encoder to the actor. Figure~\ref{aacc-hybird} shows that Hybrid Asymmetric Actor-Critic (AACC-hybrid) adds different Env Encoders to the actor and the critic respectively.

\begin{figure}[h]
\centering
\subfigure[Robust]{
\includegraphics[width=7cm]{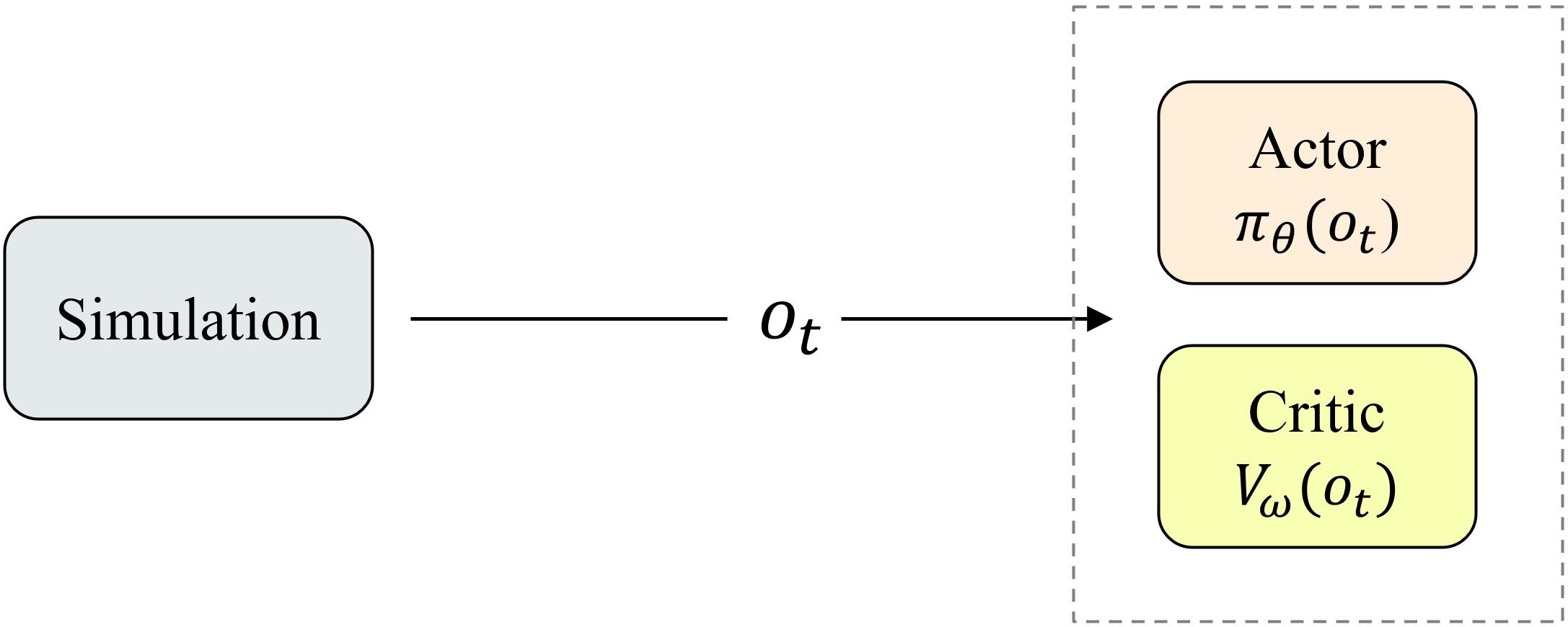}
\label{robust}
}
\quad
\subfigure[SysID]{
\includegraphics[width=7cm]{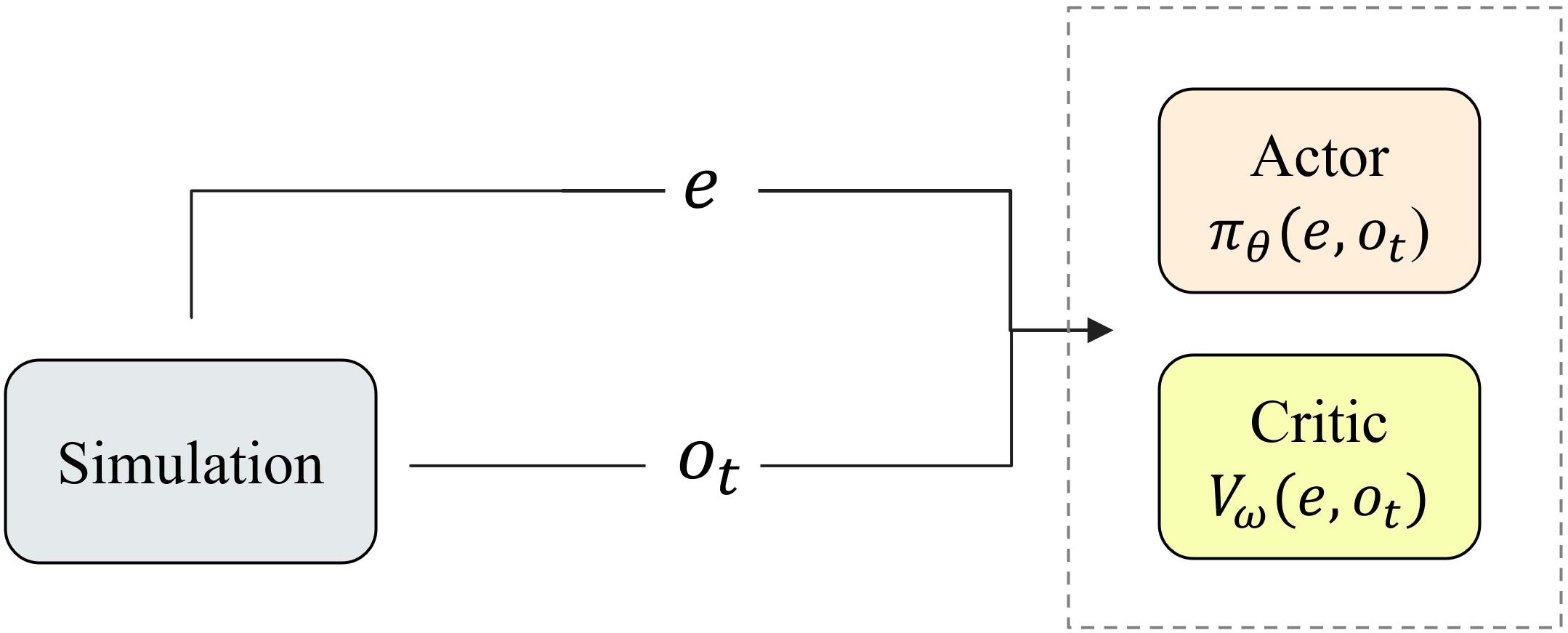}
\label{sysid}
}

\vspace{.5in}

\subfigure[RMA]{
\includegraphics[width=7cm]{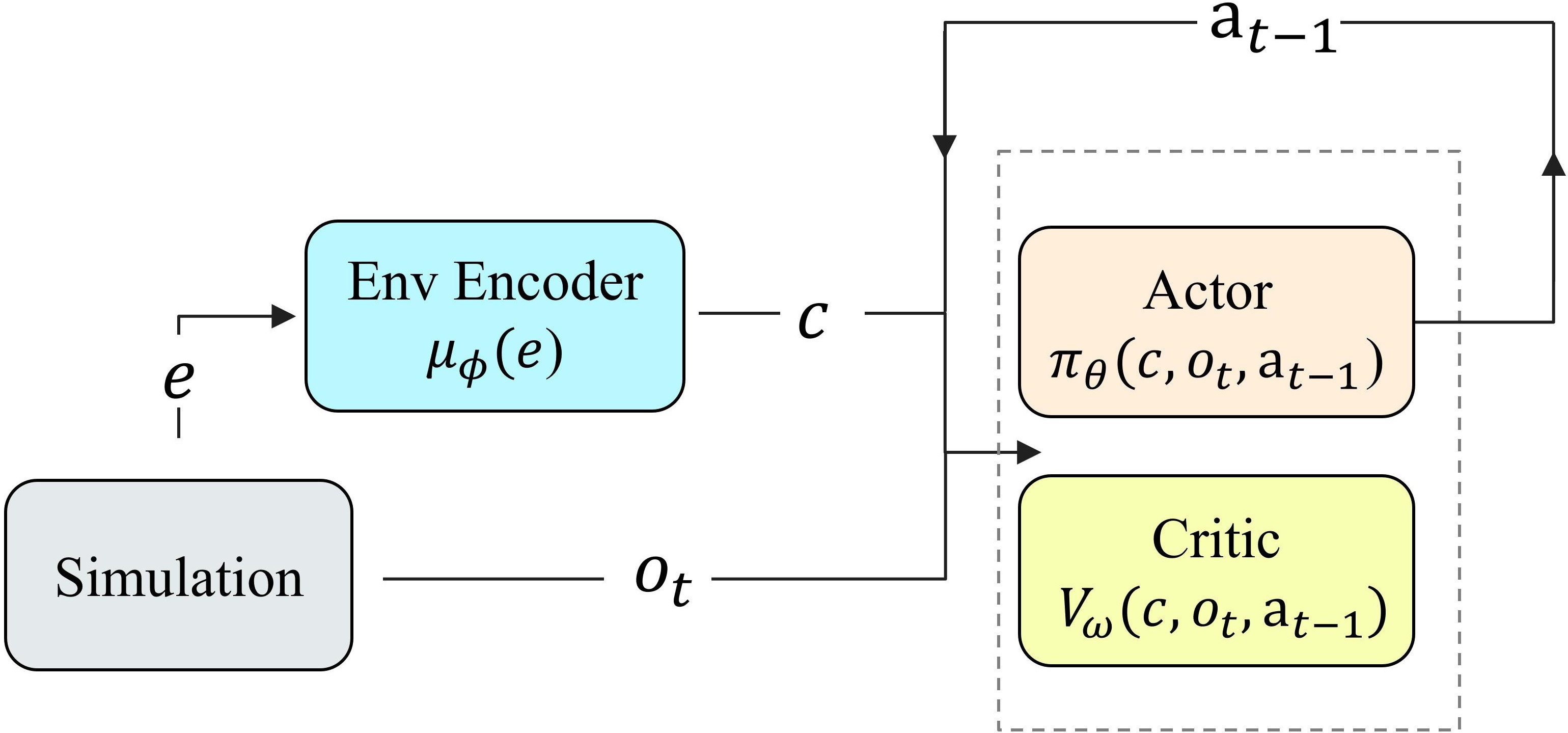}
\label{rma}
}
\quad
\subfigure[RMA-normal]{
\includegraphics[width=7cm]{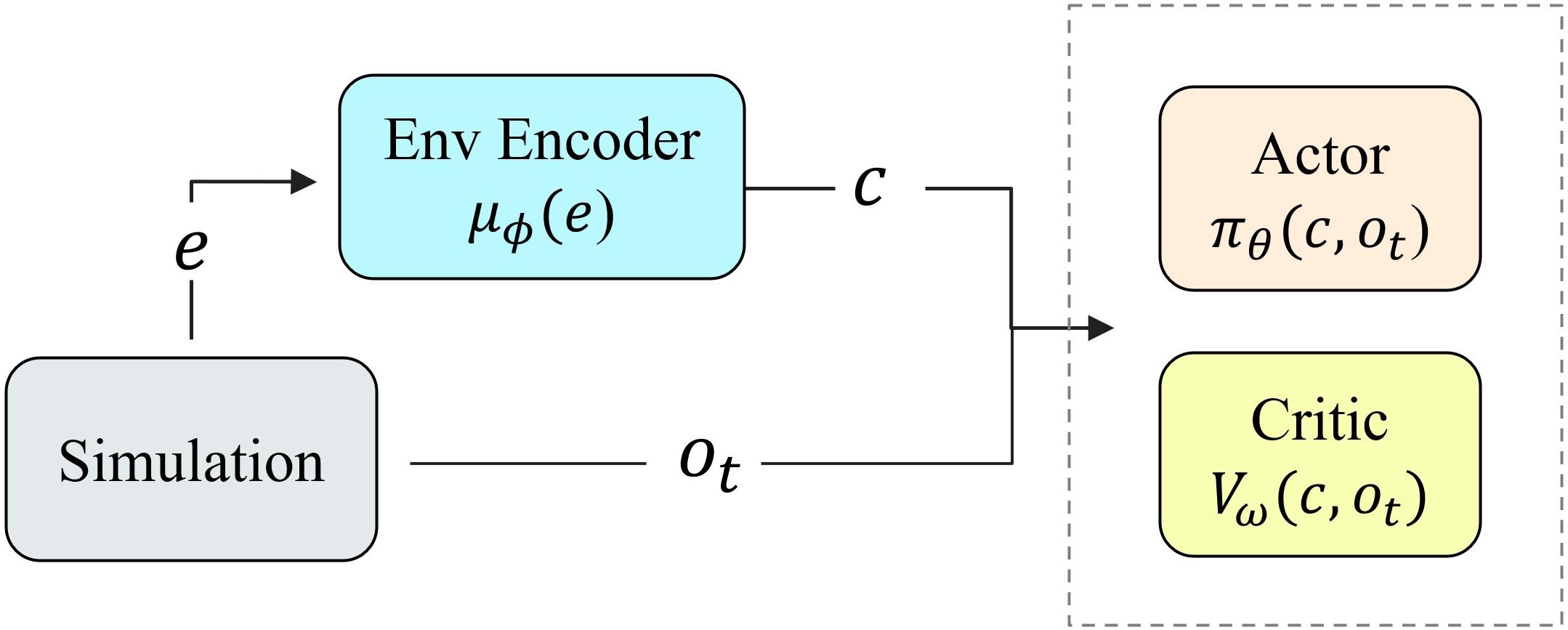}
\label{rma-normal}
}

\vspace{.5in}

\subfigure[AACC-actor]{
\includegraphics[width=7cm]{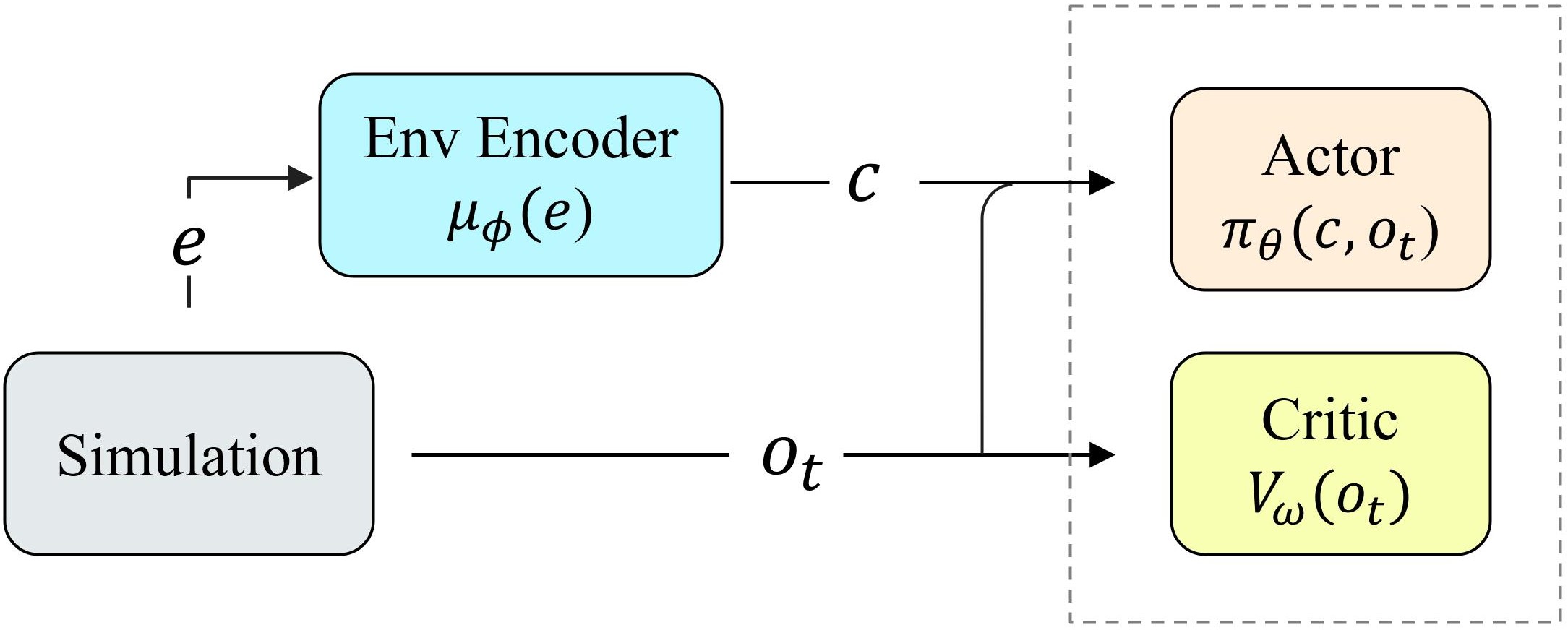}
\label{aacc-actor}
}
\quad
\subfigure[AACC-hybrid]{
\includegraphics[width=7cm]{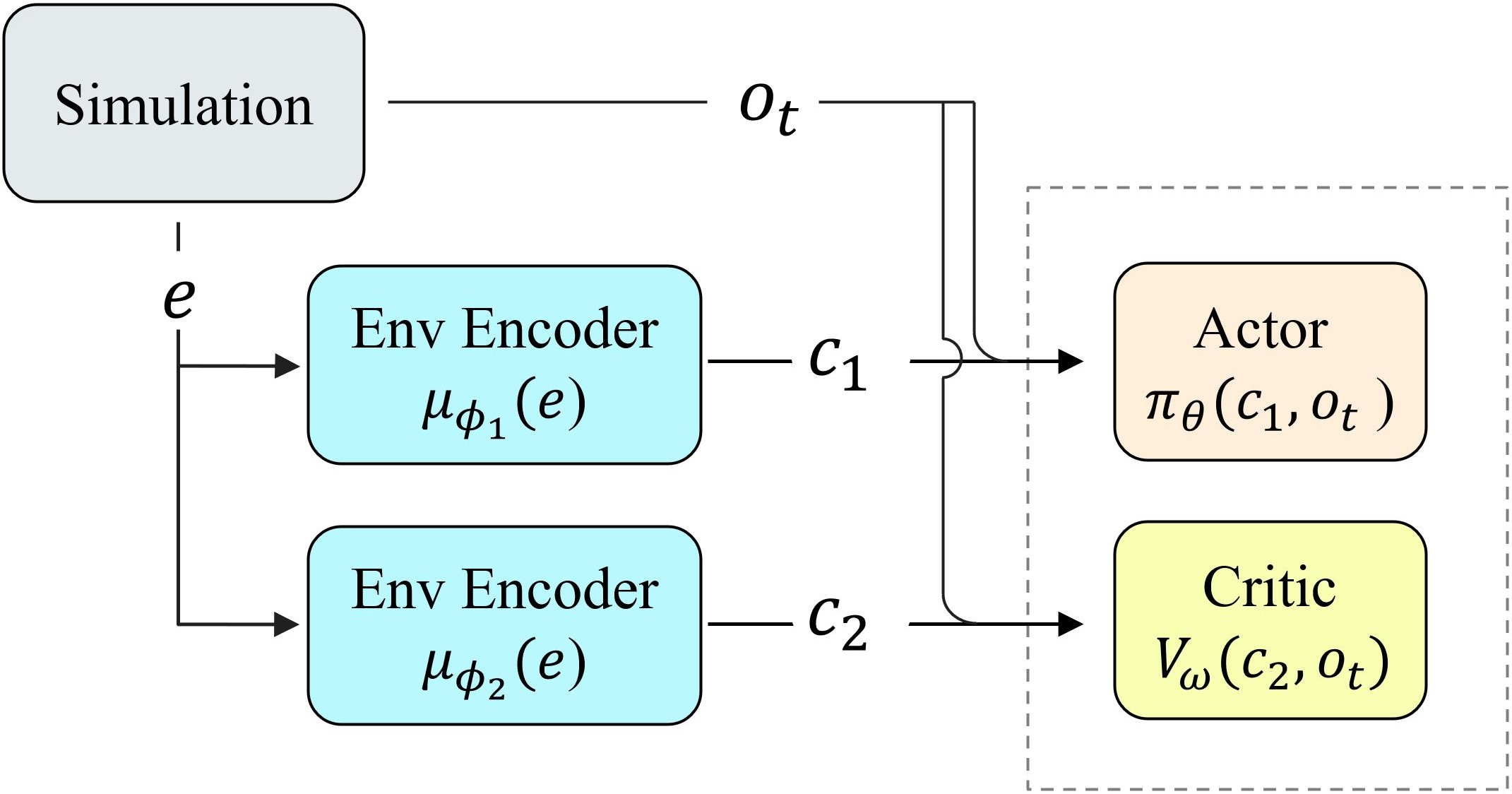}
\label{aacc-hybird}
}
\caption{The framework of six baseline methods.}
\end{figure}

\newpage

\section{Additional Experiments}\label{additional experiment}
This section includes some additional experiments and these results. 
Figure~\ref{without encoder} shows the evaluation curves of AACC and AACC without Env Encoder in Gym-JSBSim, Hopper, and Laikago.
Figure~\ref{Different Distribution} shows the evaluation curves that the environmental factors during evaluation follow new distributions.
Figure~\ref{Unseen Scenarios} shows the evaluation curves that the environmental factors are slightly shifted during evaluation.
Figure~\ref{dimension of actor} shows the evaluation curves of the varying the dimension of the output of Env Encoder to the actor of AACC-hybrid.
Figure~\ref{env factor} shows how different environmental factors affect the evaluation curve. Down wind curve (purple) converges the fastest and all environmental factors curve converges faster than Robust.
Figure~\ref{heat map} shows the performance of a trained AACC in different cases of environmental factors. 
Figure~\ref{adaptation} shows a UAV agent can adapt to changing wind environment.

\subsection{Environmental Factors} \label{D1}
To further evaluate the effect of each environmental factor on the training process, we pick up one certain environmental factor as the input of the critic network to train AACC without Env Encoder, while other environmental factors keep on changing during each training episode.
The reason we choose AACC without Env Encoder is that there is just one environmental factor, so there is no need to add the environmental factor encoder. In Gym-JSBSim, the environmental factors include north wind, east wind and down wind. And in Figure~\ref{heat map}, we use a heat map to depict the performance of AACC for various environmental factors of Gym-JSBSim. The result shows that down wind has the greatest impact on the performance of AACC. We use each environmental factor to train AACC without Env encoder respectively. In Figure~\ref{env factor}, we observe down wind has the most influence on training speed. We can infer that the degree of acceleration of training depends on the effect of the environmental factor on performance.

\subsection{Continuous Adaptation}
All the experiment settings assume that the environmental factors do not change within an episode. To allow AACC to respond to a constantly changing scenario, we relax this assumption by re-sampling the testing environmental factors within an episode with a low probability. We take an experiment in Gym-JSBSim that the wind speed and the wind direction will change within an episode. The result is shown in Figure~\ref{adaptation} that the success ratio of steady flight is 90\%. It shows that the UAV agent is still able to steady flight when wind speed and wind direction change within an episode, although the agent is trained with fixed environmental factors within an episode.

\begin{figure}[!t]
\centering
\subfigure[Gym-JSBSim]{
\includegraphics[width=5cm]{figures/ablation2.png}
}
\quad
\subfigure[Hopper]{
\includegraphics[width=5cm]{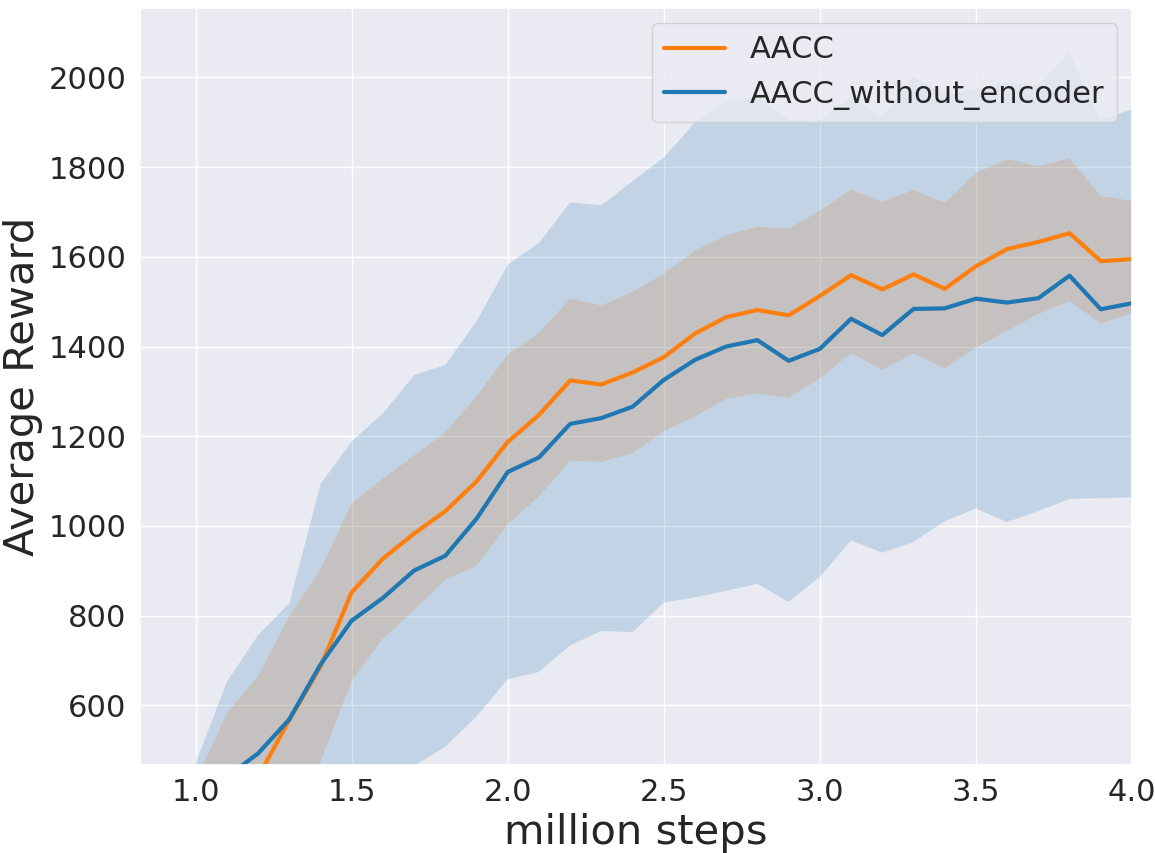}
}
\quad
\subfigure[Laikago]{
\includegraphics[width=5cm]{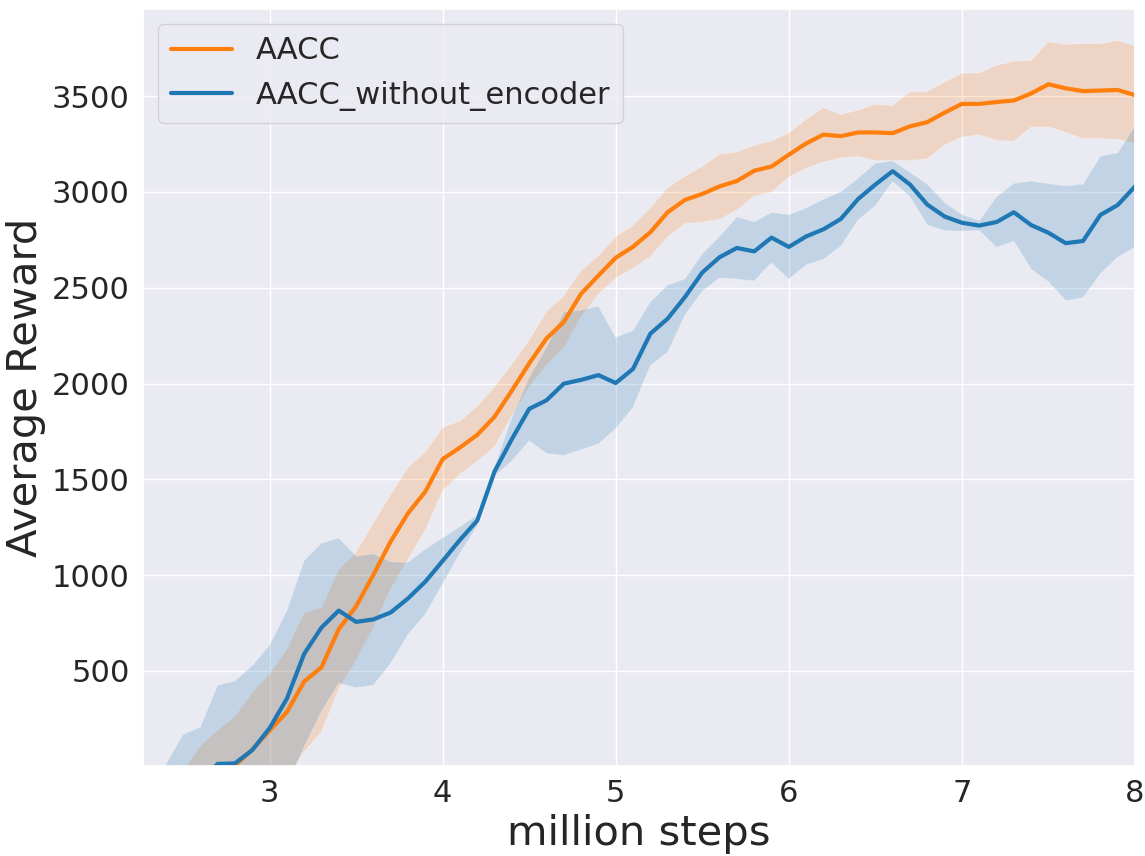}
}
\caption{With vs. Without Env Encoder}
\label{without encoder}
\end{figure}

\begin{figure}[!t]
\centering
\subfigure[Gym-JSBSim]{
\includegraphics[width=5cm]{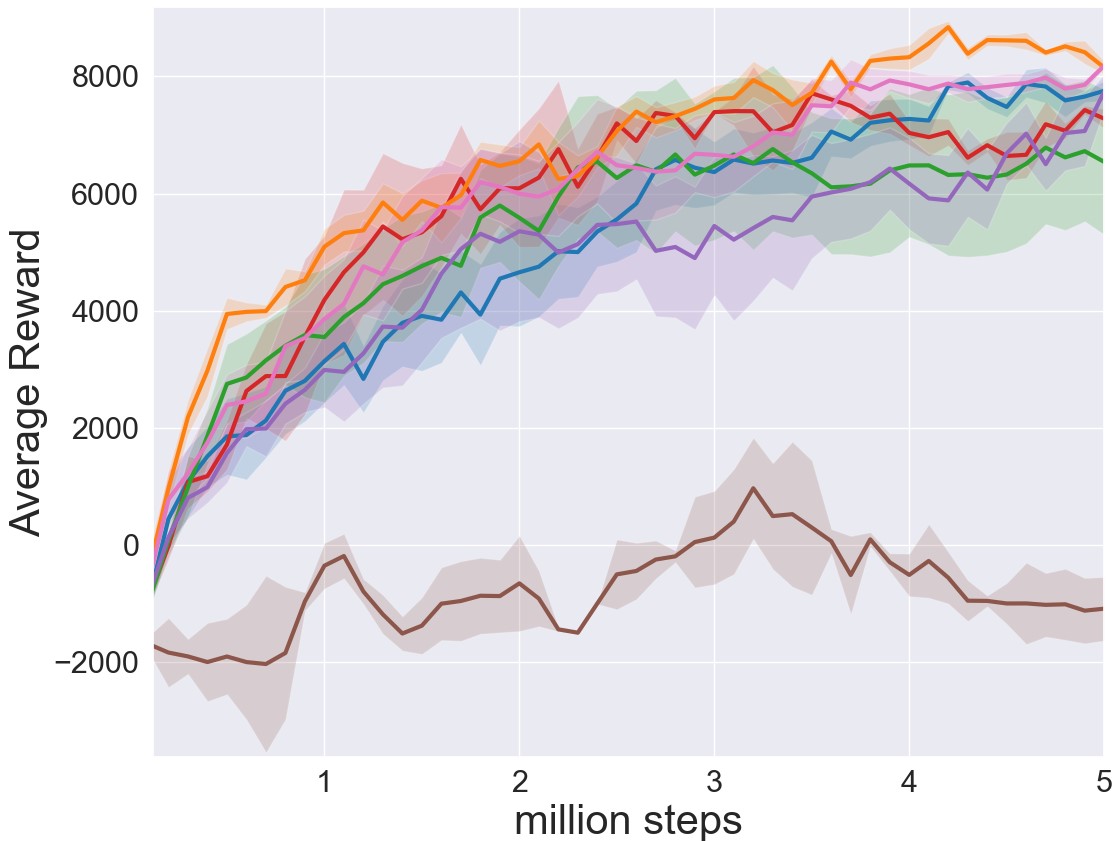}
}
\quad
\subfigure[Laikago]{
\includegraphics[width=5cm]{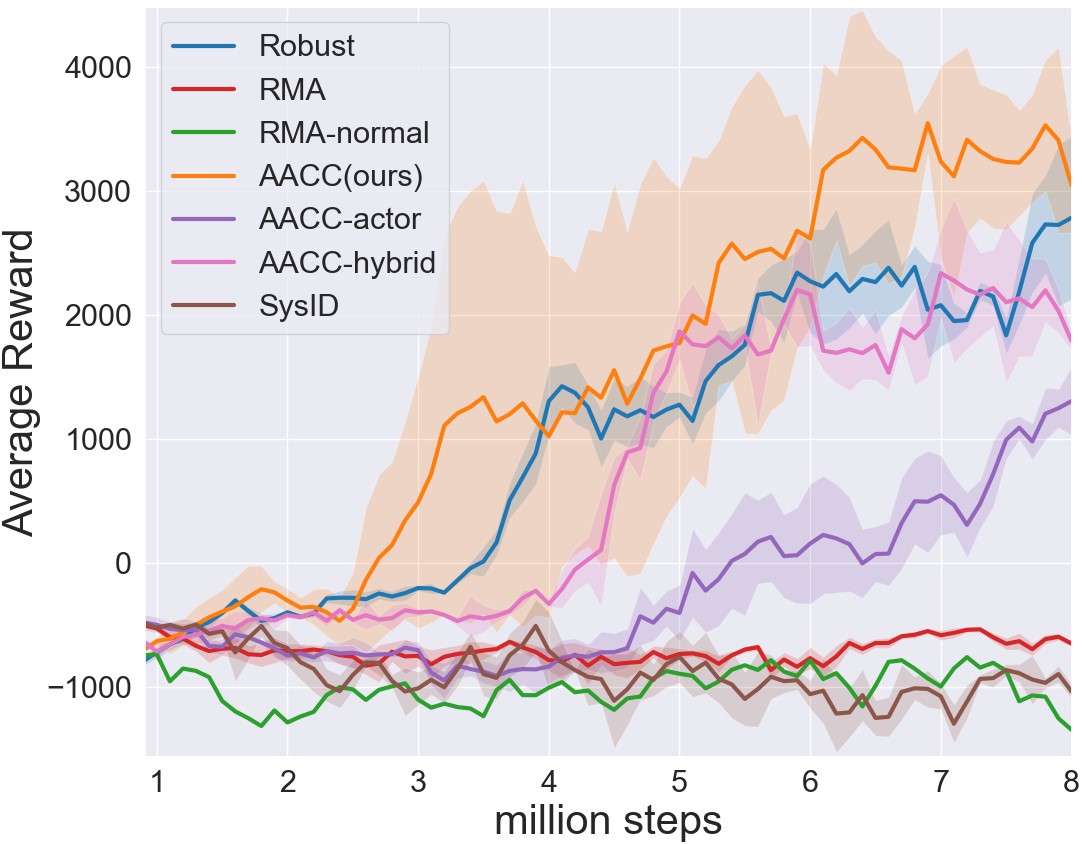}
}
\caption{Different Distribution}
\label{Different Distribution}
\end{figure}

\begin{figure}[!t]
\centering
\subfigure[Gym-JSBSim]{
\includegraphics[width=5cm]{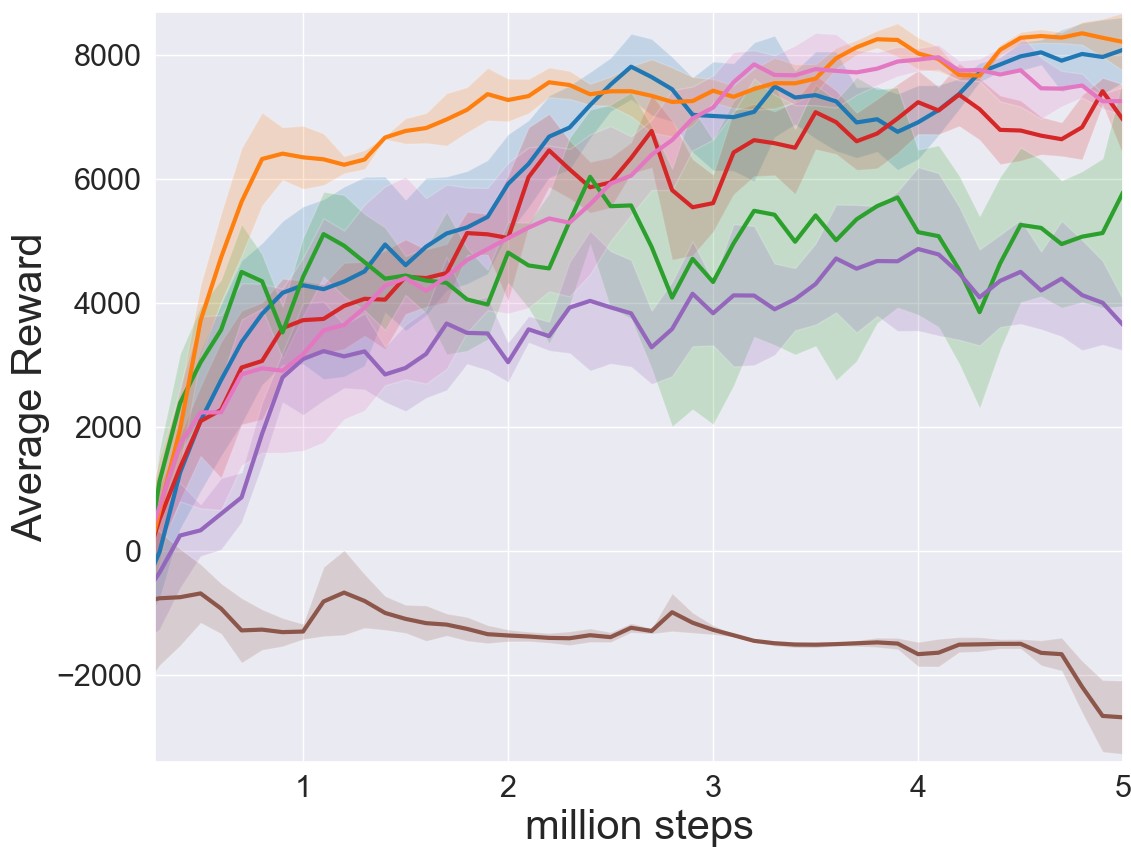}
}
\quad
\subfigure[Laikago]{
\includegraphics[width=5cm]{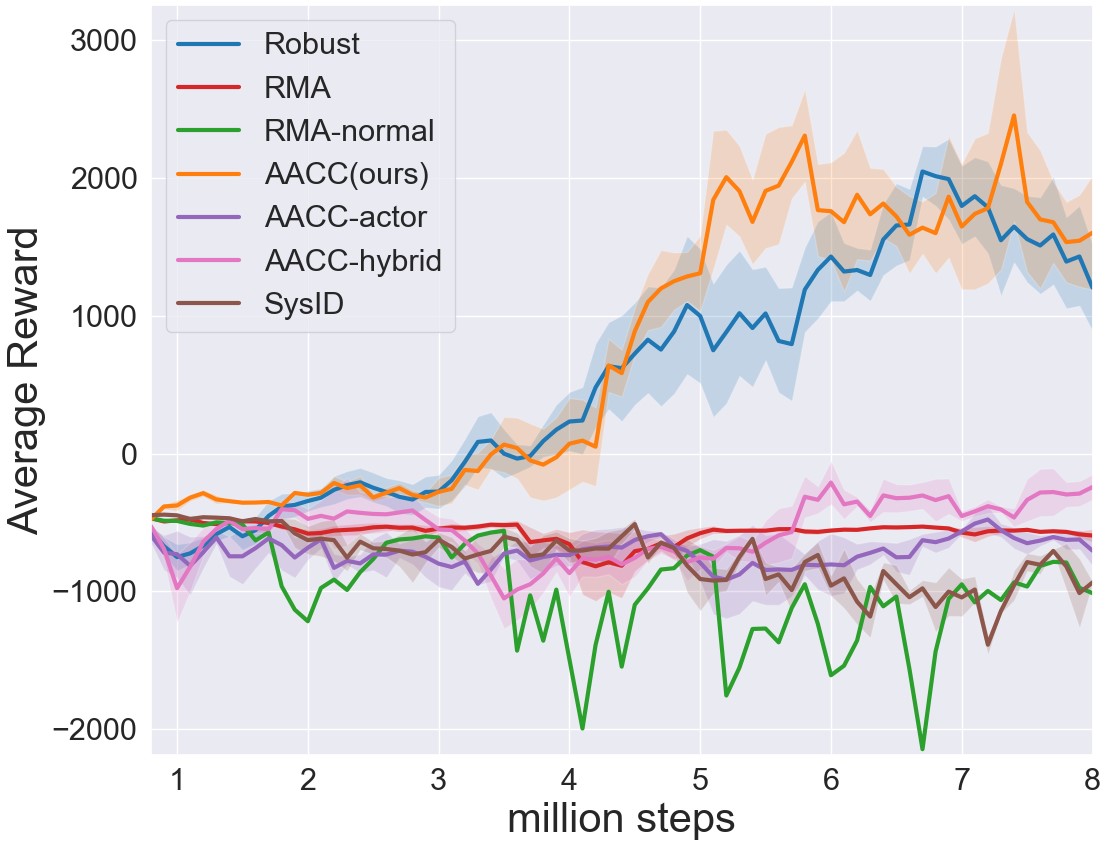}
}
\caption{Unseen Scenarios}
\label{Unseen Scenarios}
\end{figure}

\begin{figure}[!t]
  \centering
\includegraphics[width=7cm]{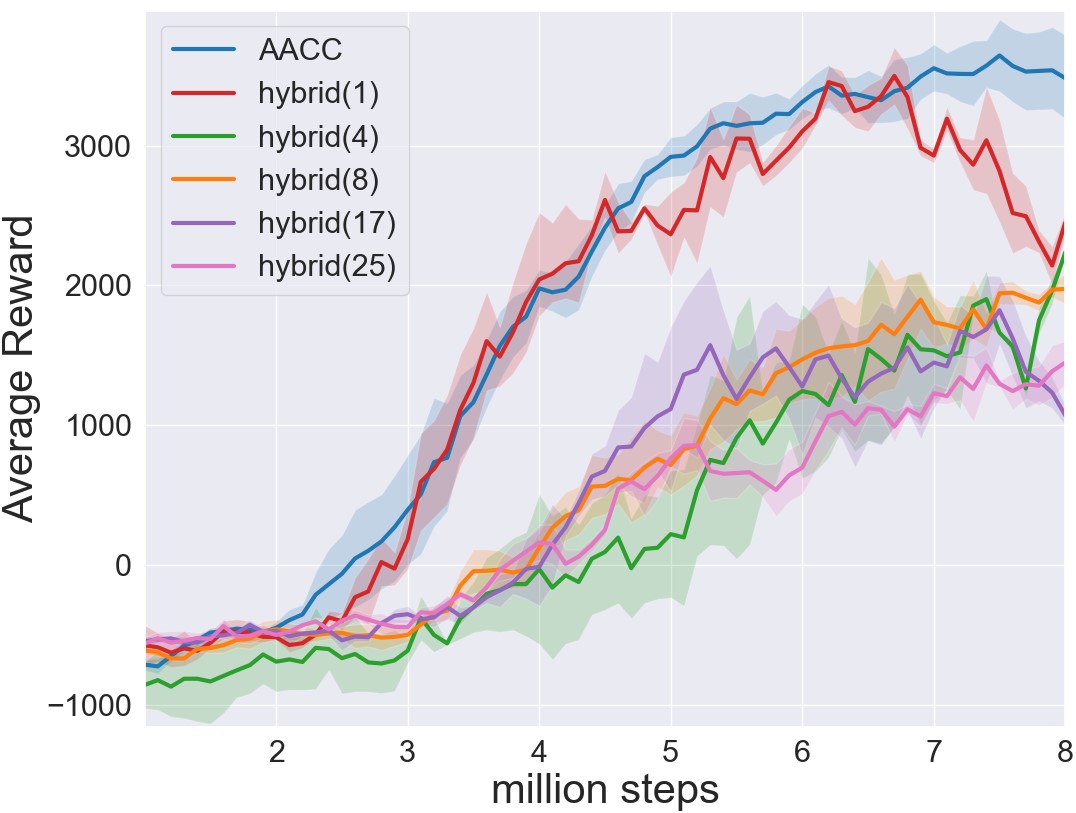}
\caption{The figure shows the evaluation curves of the varying the dimension of the output of Env Encoder to the actor of AACC-hybrid.}
\label{dimension of actor}
\end{figure}

\begin{figure}[!t]
  \centering
\includegraphics[width=7cm]{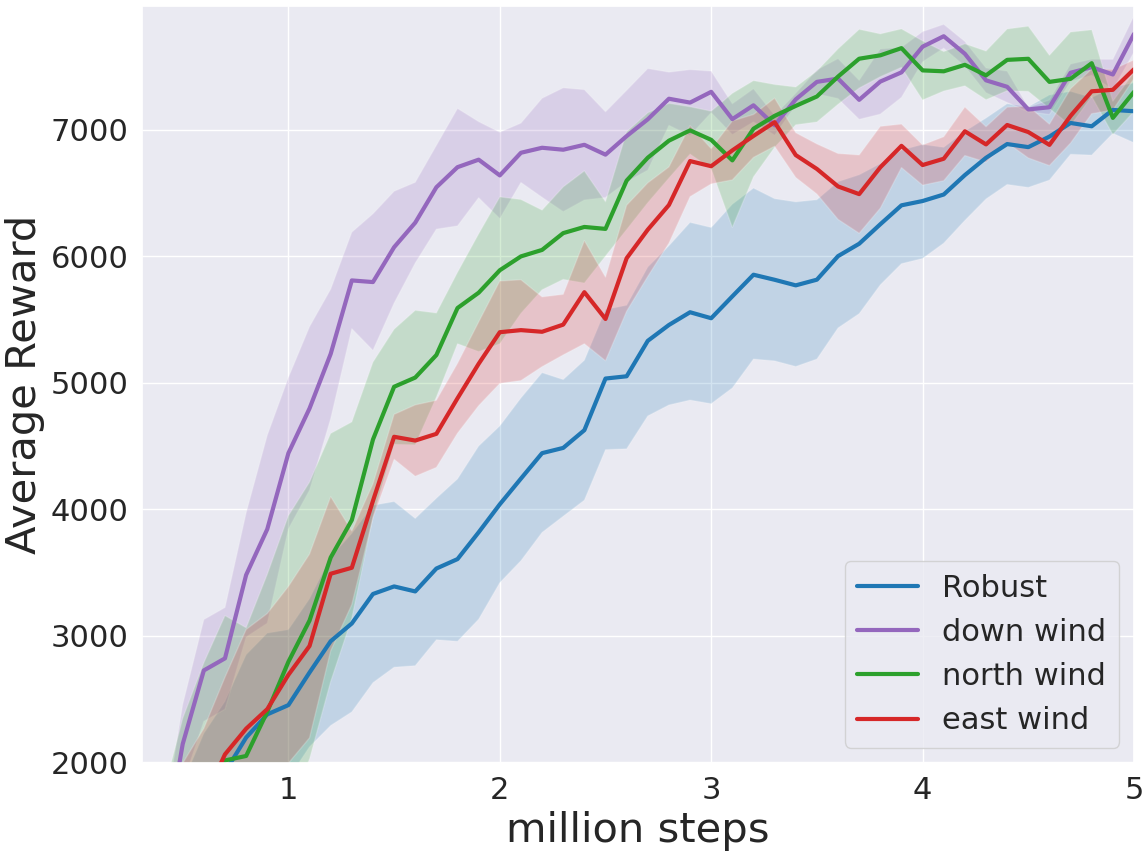}
\caption{The figure shows how different environmental factors affect evaluation curve. Down wind curve (purple) converges the fastest and all environmental factors curve converge faster than the Robust baseline.}
\label{env factor}
\end{figure}

\begin{figure}[!t]
  \centering
  \includegraphics[width=12cm]{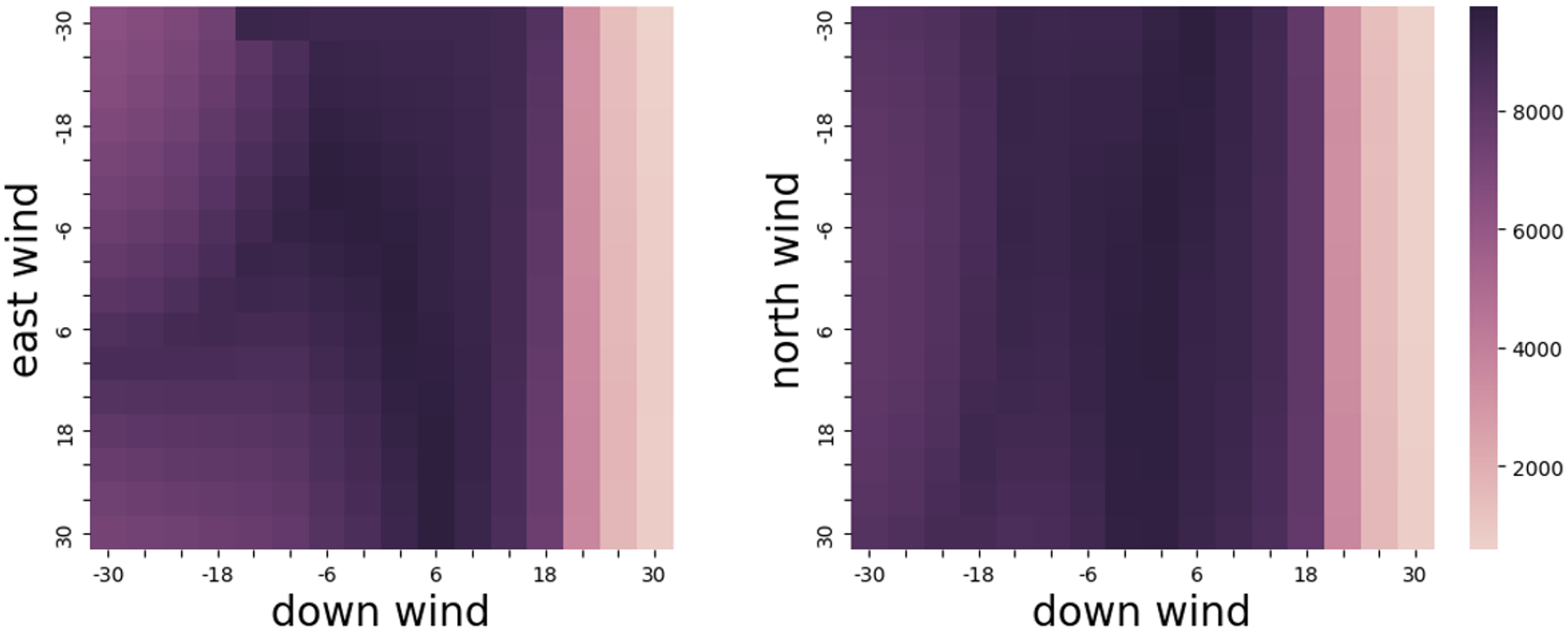}
  \caption{Heat Map.}
  \label{heat map}
\end{figure}

\begin{figure}[!t]
  \centering
  \includegraphics[width=12cm]{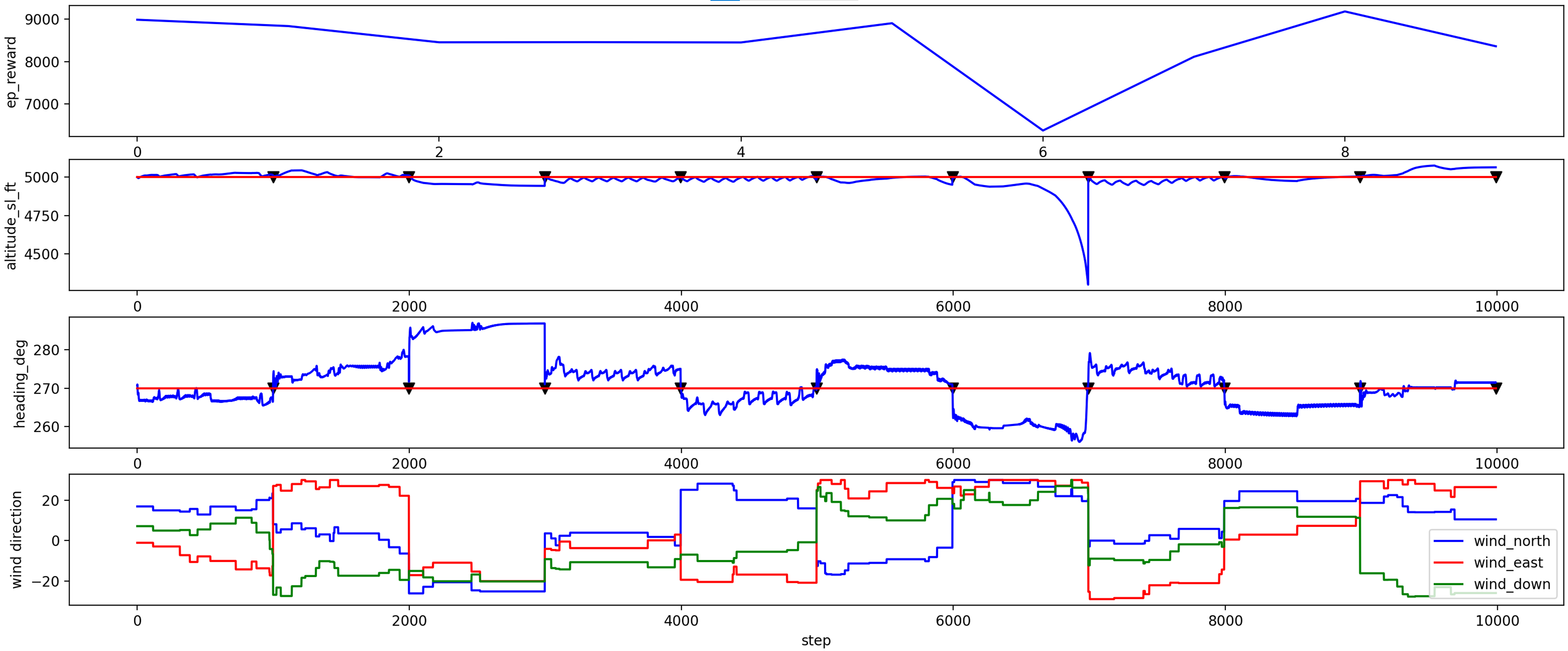}
  \caption{Continuous Adaptation.}
  \label{adaptation}
\end{figure}

\newpage

\section{Hyperparameters} \label{hyperparameters}
Table~\ref{hyperparameter} lists the common hyperparameters AACC using PPO in the comparative evaluation. Table~\ref{output dimension} lists the dimension of the output of the environmental factor encoder in each environment.

\begin{table}[t]
    \centering
    \caption{AACC using PPO hyperparameters}
    \begin{tabular}{l|r}
      \toprule
      \bfseries Parameter & \bfseries Value  \\
      \midrule
      optimizer & Adam \\
      learning rate of actor network & 3e-4 \\
      learning rate of critic network & 1e-3 \\
      learning rate of environmental factor encoder & 5e-4 \\
      discount factor & 0.99 \\
      update epoch & 30 \\
      clip ratio & 0.2 \\
      batch size & 4000 \\
      nonlinearity & Tanh \\
      \bottomrule
    \end{tabular}
    \label{hyperparameter}
\end{table}

\begin{table}[t]
    \centering
    \caption{dimension of output of the environmental factor encoder}
    \begin{tabular}{l|c}
      \toprule
      \bfseries Name & \bfseries Dimension  \\
      \midrule
      Acrobot & 4 \\
      CartPole & 3 \\
      Pendulum & 2 \\ 
      Gym-JSBSim & 3 \\
      Laikago & 8 \\
      Hopper & 4 \\
      \bottomrule
    \end{tabular}
    \label{output dimension}
\end{table}

\end{onecolumn}
\end{document}